\documentclass[a4paper,fleqn]{cas-sc}
 \usepackage[numbers]{natbib}
\usepackage{amsmath,amsfonts,amssymb,amsthm}
\usepackage{mathrsfs}
\usepackage{algorithm,algorithmic}
\usepackage{array,booktabs,multirow,diagbox}
\usepackage{bm}
\usepackage{textcomp}
\usepackage{float}
\usepackage{url}
\usepackage{verbatim}
\usepackage{balance}
\usepackage{pifont}
\usepackage{color,xcolor}
\usepackage{caption}
\usepackage{subcaption}
\usepackage{graphicx}
\usepackage{hyperref}

\makeatletter
\newtheorem{remark}{Remark}

\begin{document}

% \shorttitle{<short title of the paper for running head>} 
%Highlights-----------
%\shorttitle{A Hybrid Search Mode-Based Differential Evolution Algorithm for Auto Design of the Interval Type-2 Fuzzy Logic System}    

% \shortauthors{<short author list for running head>}
%\shortauthors{Shunze Cao, Xiao Feng et al.}

\title[mode = title]{Momentum-constrained Hybrid Heuristic Trajectory Optimization Framework with Residual-enhanced DRL for Visually Impaired Scenarios}  

% \tnotemark[<tnote number>] 
\tnotemark[1]

\tnotetext[1]{This work was supported in part by the funding project under Grant ZYGX2025YGLHI001, in part by the National Natural Science Foundation of China under Grants 62276055 and 62406062, in part by the Sichuan Science and Technology Program under Grant 2023YFG0288, in part by the Natural Science Foundation of Sichuan Province under Grant 2024NSFSC1476, in part by the National Science and Technology Major Project under Grant 2022ZD0116100, in part by the Sichuan Provincial Major Science and Technology Project under Grant 2024ZDZX0012.} 
%\tnotetext[2]{The second title footnote which is a longer text matter to fill through the whole text width and overflow into another line in the footnotes area of the first page.}

\author[1]{Yuting Zeng$^{1,}$}
\ead{1030839769zengyuting@gmail.com}
\author[1]{Zhiwen Zheng$^{1,}$}
\ead{z.z.w@icloud.com}
\author[1]{Jingya Wang}
\ead{jingya.wang@std.uestc.edu.cn}
\author[1]{You Zhou}
\ead{1244568901@qq.com}
\author[2]{JiaLing Xiao}
\ead{jollyxiao1998@163.com}
\author[1]{Yongbin Yu}
\ead{ybyu@uestc.edu.cn}
\author[1]{Manping Fan$^{*,}$}
\ead{fmpfmp@uestc.edu.cn}
\author[2,3]{Bo Gong}
\ead{gongbo@med.uestc.edu.cn}
\author[1]{Liyong Ren}
\ead{lyren@uestc.edu.cn}

\address[1]{School of Information and Software Engineering, University of Electronic Science and Technology of China, Chengdu 610054, Sichuan, P.R. China}
\address[2]{Sichuan Provincial Key Laboratory for Human Disease Gene Study, Sichuan Academy of Medical Sciences \& Sichuan Provincial People's Hospital, University of Electronic Science and Technology of China, Chengdu, Sichuan, China.}
\address[3]{Research Unit for Blindness Prevention, Chinese Academy of Medical Sciences, Sichuan Academy of Medical Sciences and Sichuan Provincial People’s Hospital, Chengdu, Sichuan, China.}

\cortext[cor1]{Corresponding authors}
\fntext[equal]{The two authors contribute equally to this work.}

\begin{abstract}
Safe and efficient assistive planning for visually impaired scenarios remains challenging, since existing methods struggle with multi-objective optimization, generalization, and interpretability. In response, this paper proposes a Momentum-Constrained Hybrid Heuristic Trajectory Optimization Framework (MHHTOF). To balance multiple objectives of comfort and safety, the framework designs a Heuristic Trajectory Sampling Cluster (HTSC) with a Momentum-Constrained Trajectory Optimization (MTO), which suppresses abrupt velocity and acceleration changes. In addition, a novel residual-enhanced deep reinforcement learning (DRL) module refines candidate trajectories, advancing temporal modeling and policy generalization. Finally, a dual-stage cost modeling mechanism (DCMM) is introduced to regulate optimization, where costs in the Frenet space ensure consistency, and reward-driven adaptive weights in the Cartesian space integrate user preferences for interpretability and user-centric decision-making. Experimental results show that the proposed framework converges in nearly half the iterations of baselines and achieves lower and more stable costs. In complex dynamic scenarios, MHHTOF further demonstrates stable velocity and acceleration curves with reduced risk, confirming its advantages in robustness, safety, and efficiency.
\end{abstract}
\maketitle

\section{Introduction}
Visually impaired individuals face significant challenges in navigating complex environments due to their limited access to visual cues, such as obstacles, road boundaries, and dynamic agents \cite{16}. In both indoor and outdoor settings, the absence of reliable visual feedback not only increases the risk of collisions and disorientation but also imposes a high cognitive load when interpreting spatial layouts through auditory \cite{9} or tactile feedback. While conventional assistive tools such as guide dogs \cite{62} and white canes offer basic support, they often fall short in structured environments with low-speed dynamic obstacles. These limitations highlight the need for intelligent agent-based assistive systems capable of providing safe, smooth, and context-aware motion guidance in real time \cite{11}, where context-awareness entails perceiving environmental semantics such as road topology, trajectory intent, and spatial affordances, together with user intent and localized spatial structures, rather than relying solely on geometric or kinematic cues.

Within agent-based assistive systems, trajectory planning for visually impaired individuals involves challenges beyond general navigation, requiring agents to operate under sensory limitations \cite{74} while simultaneously ensuring interpretability and user comfort. This calls for planning frameworks that are not only robust and semantically coherent, but also aligned with human perception and behavior \cite{75}, so that motion decisions remain safe, smooth, and perceptually consistent with users’ sensory channels.

However, meeting these requirements remains challenging for existing trajectory planning methods. Sampling-based approaches \cite{77} allow rapid generation of candidate trajectories through polynomial parameterizations in the Frenet frame and can efficiently explore the planning space. Yet, they are often constrained to static environments and struggle to balance safety, comfort, and smoothness in dynamic settings. In particular, they lack effective mechanisms to handle momentum continuity and dynamic feasibility, leading to abrupt velocity or acceleration variations. To address these limitations, this work introduces MTO, where third-order interpolation combined with momentum continuity constraints produces candidate trajectories that are not only physically feasible but also aligned with human comfort. This design explicitly suppresses abrupt velocity and acceleration variations, thereby improving both the smoothness and dynamic feasibility of the trajectories.

Recent advancements in DRL have been applied to diverse navigation tasks. For instance, DRL has been employed in the Frenet for lane-change planning with grid-map inputs \cite{71}. Comparative evaluations \cite{63} show that Deep Deterministic Policy Gradient (DDPG) outperforms Proximal Policy Optimization (PPO) in convergence speed and reward accumulation for driving tasks, yet lacks user-specific safety modeling. Other efforts modify PPO, such as Beta-distribution-based policies with distributed sampling for ship navigation \cite{64}, yielding smoother trajectories and improved robustness but still omitting trajectory priors and semantic constraints. Overall, DRL approaches often suffer from unstable generalization, limited temporal modeling, and weak interpretability, which restrict their applicability in safety-critical assistive scenarios for visually impaired users. Inspired by ResRace \cite{69} and RSAC \cite{70}, we incorporate residual learning to refine policy outputs using prior-guided baselines, thereby accelerating convergence and reducing data inefficiency. In parallel, temporal dependencies are captured through long short-term memory (LSTM), which have proven effective in maritime and mobile robot tasks \cite{60,61}, enabling robust sequential decision-making. To this end, we design a residual-enhanced, temporal-based Actor–Critic network that strengthens feature representation, improves training stability, and enhances temporal coherence in trajectory decisions.

Hybrid approaches have attempted to combine sampling and learning \cite{6,7}, yet they often provide insufficient multi-objective balancing and lack human-centered optimization mechanisms, making it difficult to meet the comfort and personalization needs of visually impaired users. Existing trajectory optimization methods typically rely on minimizing handcrafted cost functions under motion constraints, which further limits adaptability and interpretability. Meta-reviews \cite{12,14} also highlight persistent deficiencies in assistive planning systems, including inadequate environmental modeling, insufficient support for multi-objective balancing \cite{76}, and limited adaptability to diverse user preferences and cognitive traits. Although some physical interaction frameworks \cite{13} show promise for real-world deployment, their dependence on hardware restricts scalability and excludes algorithmic trajectory optimization. To overcome these issues, we propose DCMM, in which heuristic optimization in the Frenet frame ensures feasibility and comfort, while reward-based refinement in the Cartesian frame incorporates personalization and interpretability through a weight-transfer scheme.

To address the limitations in multi-objective optimization, generalization, and interpretability, we propose an MHHTOF specifically tailored for visually impaired scenarios. The framework integrates structured global Heuristic Trajectory Sampling with Multi-Objective Evaluation (HTSCMOE) integrating with a residual-enhanced DRL policy, achieving a tight coupling between deterministic efficiency and adaptive robustness. The main contributions of this paper are summarized as follows:
\begin{enumerate}[1)]
\item This work designs a MHHTOF that unifies heuristic sampling with residual-enhanced DRL. A shared multi-objective cost function governs both trajectory generation and policy optimization, achieving coherent and goal-aligned learning across stages.
\item To ensure trajectory feasibility and human comfort in assistive scenarios, a dynamic-aware sampling strategy is constructed using third-order interpolation and momentum continuity constraints. This approach effectively generates smooth and kinematically consistent path candidates suitable for real-world execution.
\item The study further devises a residual-enhanced actor–critic architecture equipped with LSTM-based temporal modeling, which improves policy representation capability, stabilizes training dynamics, and facilitates adaptive refinement of candidate trajectories under sequential decision-making tasks.
\item The novel human-centered DCMM with weight transfer is proposed, where heuristic sampling in the Frenet frame ensures feasibility and comfort via a multi-objective cost function, and reward-driven refinement in the Cartesian frame enhances trajectory quality. A weight transfer mechanism aligns both stages, enabling interpretable and personalized optimization for visually impaired users.
\end{enumerate}

This paper is organized as follows: Section~\ref{p} introduces the theoretical foundation and modeling assumptions; Section~\ref{h} describes the overall system architecture, including the heuristic sampling and DRL modules; Section~\ref{r} presents experimental evaluations and ablation studies; and Section~\ref{c} concludes the paper and discusses future work directions.

\section{Problem Formulation of Trajectory Optimization for Visually Impaired Scenarios}
\label{p}
This section formulates the trajectory optimization problem for visually impaired navigation. While existing Frenet-based formulations provide efficiency and analytic tractability, they lack sufficient semantic adaptability to reflect user-centric requirements such as comfort, perceived safety, and social compliance. To overcome these limitations, we introduce DCMM, which unifies geometric feasibility and semantic adaptability within a consistent framework. Serving as the backbone of the proposed planning architecture, DCMM enhances interpretability and robustness while ensuring trajectories remain both computationally tractable and cognitively aligned with real-world assistive navigation \cite{20,21}. Its two stages will be detailed in Section~\ref{h}.

\subsection{Coordinate Decoupling for Perceptual Motion Representation}
\label{p1}
The Frenet frame defines local motion along a reference path \cite{40} by projecting the agent’s position onto longitudinal and lateral directions. This allows trajectory generation directly over environment-conforming geometries such as sidewalks or pedestrian lanes \cite{22}. Compared with Cartesian coordinates, the Frenet formulation offers superior alignment with how visually impaired individuals perceive motion, as it relies more on edge-following and structured cues rather than global localization \cite{78}.

\begin{figure}
\centering
\includegraphics[width =0.45\textwidth]{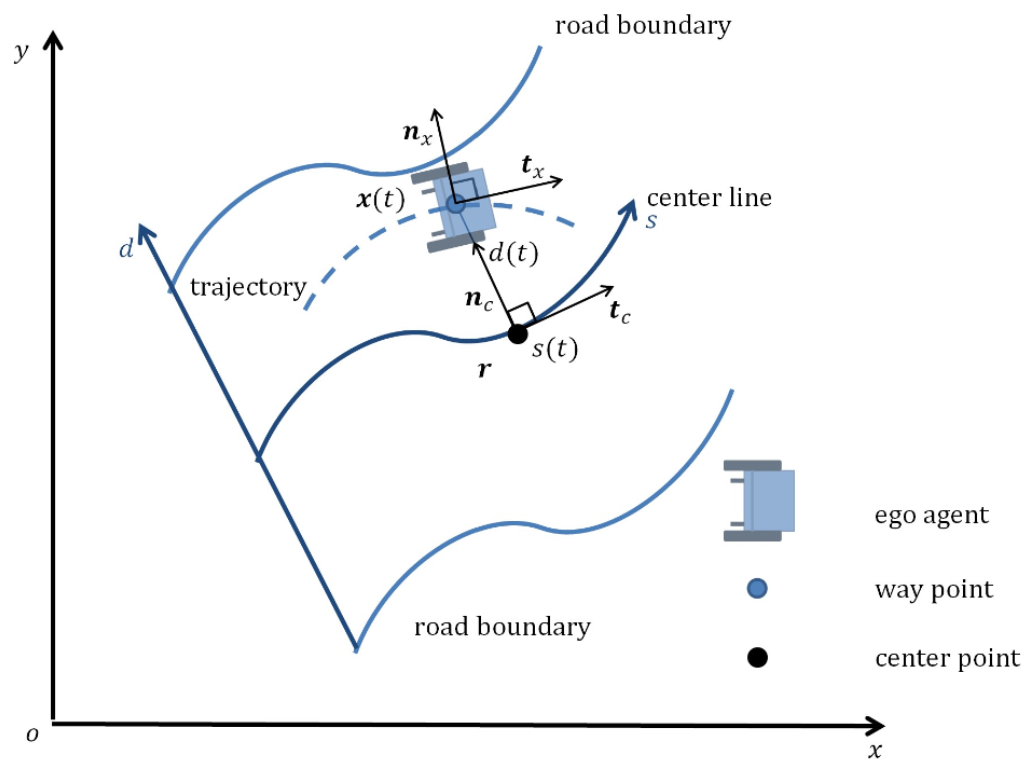}
\caption{Schematic diagram of the conversion trajectory from the Frenet coordinate to the Cartesian Coordinate.}
\label{fig1}
\end{figure}

As illustrated in Figure \ref{fig1}, the spatial state of agent is represented by its arc length $s(t)$ and lateral deviation $d(t)$ from a known reference path:
\begin{equation}
\begin{aligned}
\boldsymbol{x}(s(t), d(t))= & \boldsymbol{r}(s(t))+d(t) \boldsymbol{n}_{c}(s(t)) \\
& \left\{\begin{array}{l}
s=s(t) \\
d=d(t)
\end{array}\right.
\end{aligned}
\end{equation}
where $r(s)$ denotes the position along the reference line, and ${n}_{c}(s)$ is the corresponding normal vector. This representation introduces natural constraints aligned with sidewalk or path structures, supporting safe, semantically and perceptually meaningful navigation \cite{79}.
The full state in the Frenet frame is described as $\left[s, \dot{s}, \ddot{s} ; d, \dot{d}, \ddot{d}, d^{\prime}, d^{\prime \prime}\right]$. $\dot{s}, \ddot{s}$ represent longitudinal velocity and acceleration, while $d^{\prime}, d^{\prime \prime}$ reflect lateral variation and curvature smoothness. Such decoupling is well-suited to hierarchical planning and reflects how movement planning by visually impaired individuals is often segmented between directional advancement and lateral path adherence.
To support physical realization and control, the Frenet state can be transformed back into Cartesian coordinates through:
\begin{equation}
\boldsymbol{x}(t)=\boldsymbol{Q}(x(t), y(t))=\mathbf{F}(s, d)
\end{equation}
where
\begin{equation}
\left\{\begin{array}{l}
x_{x}(t)=\mathbf{F}_{x}(s(t), d(t)) \\
y_{x}(t)=\mathbf{F}_{y}(s(t), d(t))
\end{array}\right.
\end{equation}

These transformations guarantee geometric consistency and serve as the basis for dynamic feasibility and constraint enforcement in planning. A complete description of the transformation parameters is presented in Table \ref{tab_transformation_parameters} \cite{25}.

\begin{table*}[htbp]
\centering
\caption{Transformation Parameters from Cartesian to Frenet Coordinates}
\begin{tabular}{llll}
\toprule
\textbf{Symbol} & \textbf{Name} & \textbf{Physical Meaning / Unit} & \textbf{Coordinate System} \\
\midrule
$s$ & Frenet longitudinal coordinate & Arc length along reference line / m & Frenet \\
$\dot{s}$ & Frenet longitudinal velocity & Velocity along reference line / m/s & Frenet \\
$\ddot{s}$ & Frenet longitudinal acceleration & Change rate of longitudinal velocity / m/s\textsuperscript{2} & Frenet \\
$d$ & Frenet lateral coordinate & Offset from reference line / m & Frenet \\
$\dot{d}$ & Frenet lateral velocity & Velocity perpendicular to reference line / m/s & Frenet \\
$\ddot{d}$ & Frenet lateral acceleration & Change rate of lateral velocity / m/s\textsuperscript{2} & Frenet \\
$l'$ & First derivative of lateral offset & Derivative of lateral offset w.r.t. longitudinal position / - & Frenet \\
$l''$ & Second derivative of lateral offset & Change rate of lateral offset w.r.t. lateral velocity / 1/m & Frenet \\
$x$ & Cartesian coordinate & Vehicle position (x, y) / m & Cartesian \\
$\theta_x$ & Heading angle & Angle between vehicle heading and x-axis / rad & Cartesian \\
$k_x$ & Curvature & Curvature of trajectory / 1/m & Cartesian \\
$v_x$ & Cartesian velocity & Vehicle speed magnitude / m/s & Cartesian \\
$a_x$ & Cartesian acceleration & Change rate of vehicle speed / m/s\textsuperscript{2} & Cartesian \\
\bottomrule
\end{tabular}
\label{tab_transformation_parameters}
\end{table*}

\subsection{DCMM for Human-Centric Planning}
Building on the Frenet-based formulation in Section~\ref{p1}, we now formalize the proposed DCMM for human-centric planning. To overcome the limitations of Frenet-only costs, DCMM serves as the backbone of the framework by combining geometric feasibility with semantic adaptability. We detail its two-stage design, beginning with a decoupled cost function in the Frenet frame:
\begin{equation}
J[d, s]=J_{d}[d]+k_{s} J_{s}[s], k_{s}>0
\end{equation}
where $J_d[d]$ penalizes lateral irregularity, such as abrupt curvature changes or deviation from navigable boundaries, and $J_s[s]$ evaluates longitudinal consistency, emphasizing velocity smoothness and deceleration feasibility \cite{80}. This decoupled formulation allows interpretable control over human-centric planning priorities, such as comfort and reactivity to environmental constraints.

Let $\mathcal{T} = \{x^{(1)}(t), x^{(2)}(t), ..., x^{(N)}(t)\}$ denote the candidate trajectories generated by minimizing $J[d, s]$ and projected to Cartesian space. To incorporate context-aware and dynamic preferences in trajectory selection, a second-stage evaluation is conducted using a DRL-guided adaptive cost function:
\begin{equation}
J_{\text{eval}}^{\text{DRL}}(x^{(i)}(t)) = \sum_{j=1}^{n} \lambda_j^{\text{DRL}} \cdot \phi_j(x^{(i)}(t)), \quad x^{(i)}(t) \in \mathcal{T}
\end{equation}
where $\phi_j(\cdot)$ represents interpretable evaluation metrics in the Cartesian space, such as curvature continuity, obstacle clearance, and social compliance, and $\lambda_j^{\text{DRL}}$ denotes adaptive weights learned by the DRL policy network based on current environmental observations.

The final trajectory is selected as:
\begin{equation}
x^*(t) = \arg\min_{x^{(i)}(t) \in \mathcal{T}} J_{\text{eval}}^{\text{DRL}}(x^{(i)}(t))
\end{equation}

\begin{remark}
% This dual-stage structure enables coarse-to-fine optimization: the first stage ensures kinematic feasibility and geometric adherence to structured environments, while the second stage encodes semantic perception and context-aware preferences, thus improving robustness and human-aligned performance in visually impaired navigation \cite{27}.
% 分别论述：现有的两种坐标系上的代价函数的优势劣势，问题，6-7行
Prior works \cite{35,80,81} restricted to Frenet-frame optimization provide efficient feasibility but fail to capture semantic and social adaptability, whereas Cartesian-only learning methods \cite{68} encode rich context but often sacrifice geometric consistency. The proposed dual-stage DCMM reconciles these limitations: the first stage secures feasible and smooth motion in structured environments, while the second stage incorporates adaptive semantic evaluation.  This integration yields trajectories that are both interpretable and robust, aligning navigation outputs with the comfort, safety, and personalization demands of visually impaired users.
\end{remark}

\section{HTSCMOE with Residual-enhanced DRL for Visually Impaired Scenarios}
\label{h}
\begin{figure*}
\centerline{\includegraphics[width =.68\textwidth]{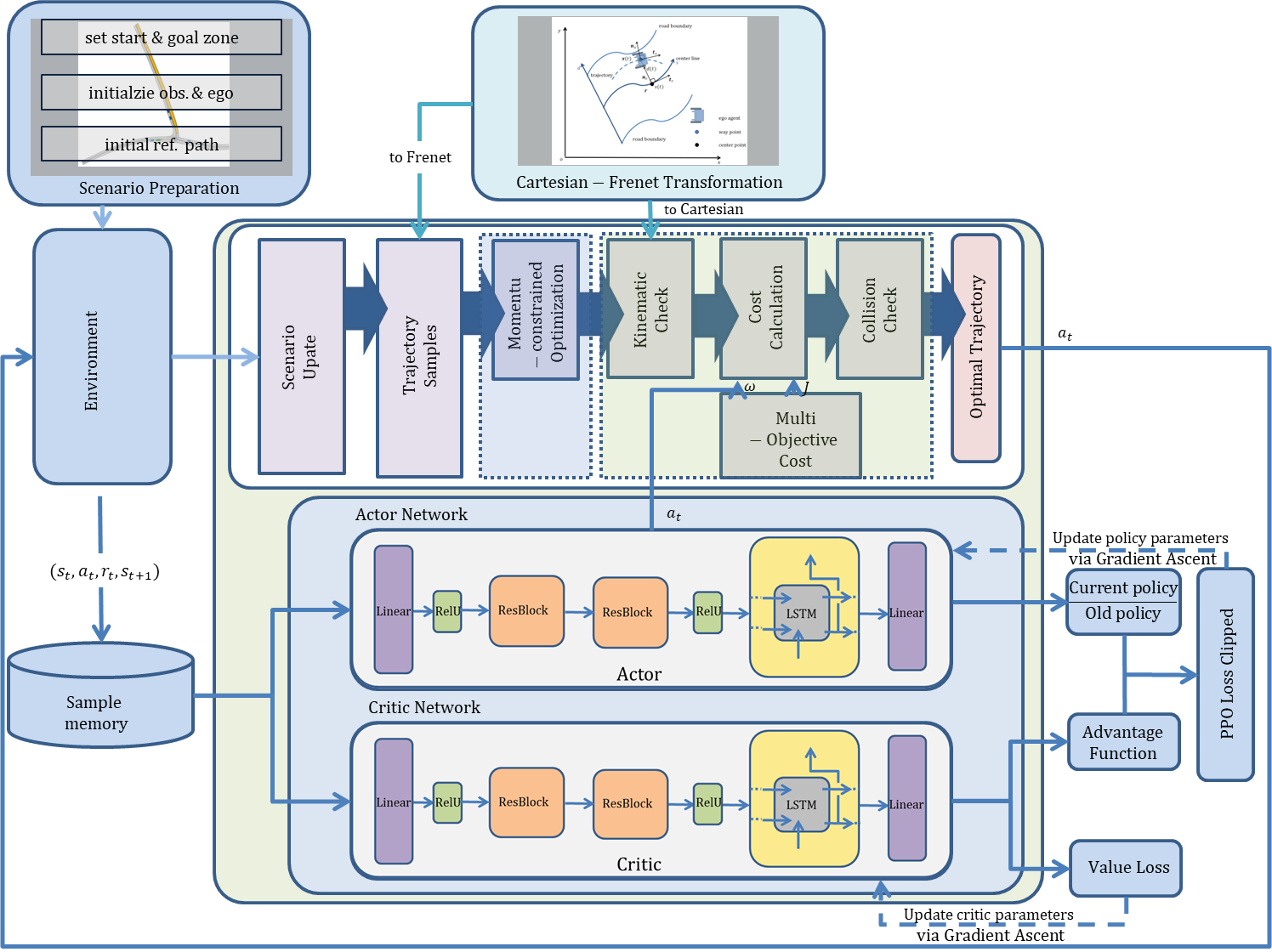}}
\caption{Overall architecture of MHHTOF.}
\label{fig2}
\end{figure*}

This chapter presents the MHHTOF framework for visually impaired, designed to achieve safe, efficient, and interpretable trajectory optimization, which integrates HTSCMOE for trajectory generation with a residual-enhanced actor–critic DRL network \cite{82}. As shown in Fig.~\ref{fig2}, the two modules interact through DCMM, which serves as the bridging cost mechanism. The following subsections detail the HTSCMOE front-end and the residual-enhanced DRL back-end with DCMM implementation.

\subsection{Design of HTSCMOE}
\begin{figure*}
\centerline{\includegraphics[width =.65\textwidth]{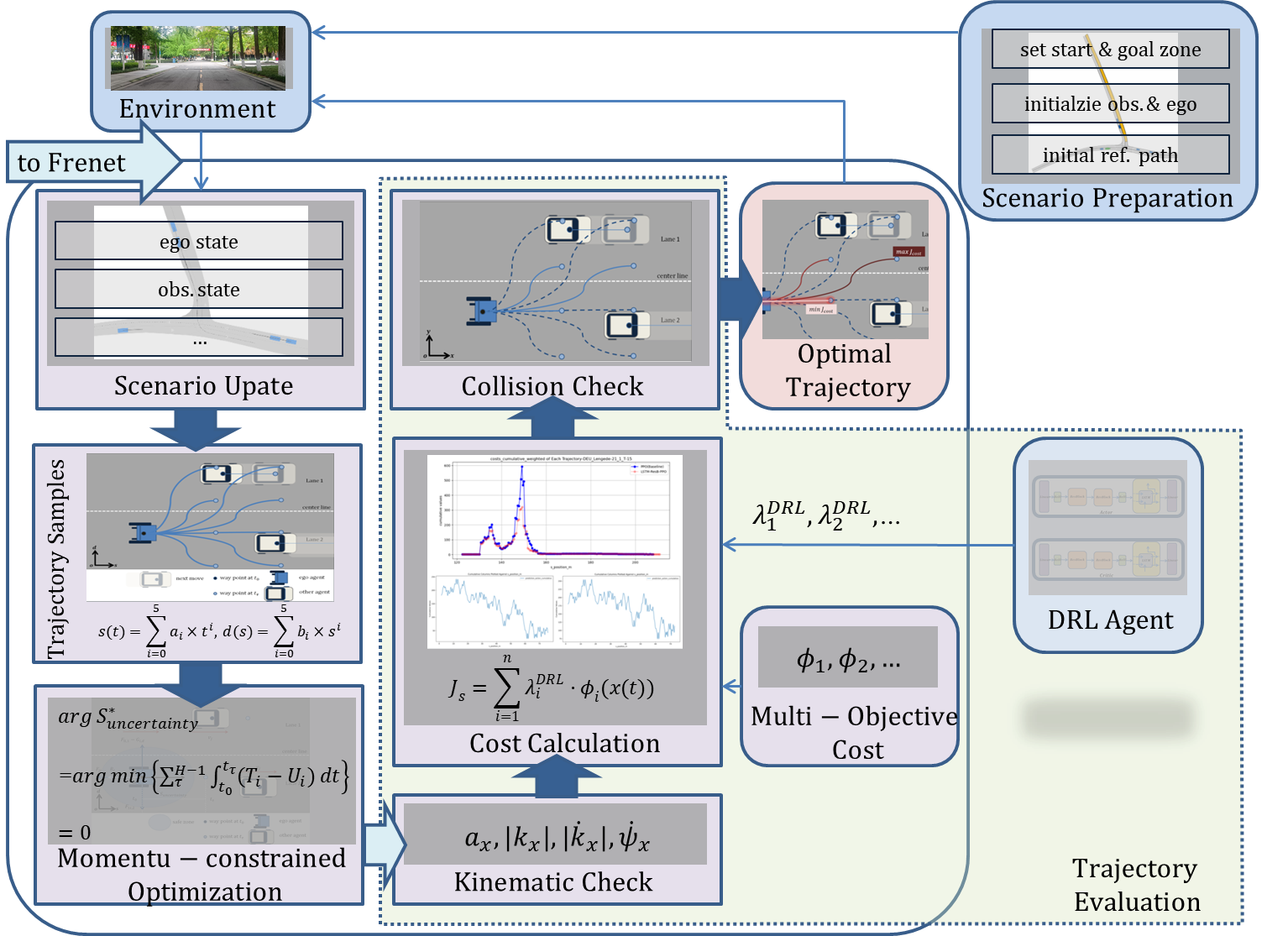}}
\caption{Flowchart of HTSCMOE.}
\label{fig3}
\end{figure*}

As shown in Fig.~\ref{fig3}, the front-end module HTSCMOE generates a compact set of safe and interpretable trajectory candidates for the DRL back-end. The process consists of HTSC Generation with endpoint smoothing, MTO, and trajectory evaluation through the first-stage DCMM. 
\subsubsection{HTSC Generation with Endpoint Smoothing for Visually Impaired}
\label{h11}
The first step generates a trajectory cluster in the Frenet frame by fitting quintic polynomials under boundary constraints, enabling efficient decoupling of longitudinal and lateral motions. To ensure temporal continuity and avoid abrupt velocity or acceleration changes, endpoint smoothing via cubic interpolation is applied, improving comfort and stability for MTO.

Each local trajectory segment within the heuristic sampling process is represented by fifth-order polynomials \cite{83} in the Frenet coordinate system. The longitudinal and lateral motions are decoupled and parameterized respectively as:
\begin{equation}
\left\{\begin{array}{l}
s(t)=\sum_{i=0}^{5} a_{i} \times t^{i} \\
d(s)=\sum_{i=0}^{5} b_{i} \times s^{i}
\end{array}\right.
\end{equation}
$s(t)$ denotes longitudinal evolution over time, while $d(s)$ captures lateral deviation with respect to the longitudinal displacement. This representation maintains coupling expressivity while enabling efficient computation.

Given the boundary state at both the start and end of each segment, the coefficients $a_i$ and $b_i$ are computed analytically:

For longitudinal motion:
\begin{equation}
\begin{array}{l}
a_{[0: 2]}=\left[s\left(t_{0}\right), \dot{s}\left(t_{0}\right), \frac{1}{2} \ddot{s}\left(t_{0}\right)\right] \\
a_{[3: 5]}=\operatorname{Solve}\left(\mathbf{A}_{1}, \mathbf{B}_{1}\right)
\end{array}
\end{equation}

For lateral motion:
\begin{equation}
\begin{aligned}
b_{[0: 2]} & =\left[d\left(t_{\tau}\right), \dot{d}\left(t_{\tau}\right), \frac{1}{2} \ddot{d}\left(t_{\tau}\right)\right] \\
b_{[3: 5]} & =\operatorname{Solve}\left(\mathbf{A}_{2}, \mathbf{B}_{2}\right)
\end{aligned}
\end{equation}

\begin{equation}
\begin{array}{l}
\mathbf{A}_{1}=\left[\begin{array}{ccc}
(t_{\tau}-t_{\mathrm{0}})^{3} & (t_{\tau}-t_{\mathrm{0}})^{4} & (t_{\tau}-t_{\mathrm{0}})^{5} \\
3 \times (t_{\tau}-t_{\mathrm{0}})^{2} & 4 \times (t_{\tau}-t_{\mathrm{0}})^{3} & 5 \times (t_{\tau}-t_{\mathrm{0}})^{4} \\
6 \times (t_{\tau}-t_{\mathrm{0}}) & 12 \times (t_{\tau}-t_{\mathrm{0}})^{2} & 20 \times (t_{\tau}-t_{\mathrm{0}})^{3}
\end{array}\right] \\
\mathbf{B}_{1}=\left[\begin{array}{c}
s\left(t_{\tau}\right)-a_{0}-a_{1} \cdot (t_{\tau}-t_{\mathrm{0}})-a_{2} \cdot (t_{\tau}-t_{\mathrm{0}})^{2} \\
\dot{s}\left(t_{\tau}\right)-a_{1}-2 \times a_{2} \cdot (t_{\tau}-t_{\mathrm{0}}) \\
\ddot{s}\left(t_{\tau}\right)-2 \times a_{2}
\end{array}\right]
\end{array}
\end{equation}
\begin{equation}
\begin{array}{l}
\mathbf{A}_{2}=\left[\begin{array}{ccc}
S_{\tau0}^{3} & S_{\tau0}^{4} & S_{\tau0}^{5} \\
3 \times S_{\tau0}^{2} & 4 \times S_{\tau0}^{3} & 5 \times S_{\tau0}^{4} \\
6 \times S_{\tau0} & 12 \times S_{\tau0}^{2} & 20 \times S_{\tau0}^{3}
\end{array}\right] \\
\mathbf{B}_{2}=\left[\begin{array}{c}
d\left(t_{\tau}\right)-b_{0}-b_{1} \cdot S_{\tau0}-b_{2} \cdot S_{\tau0}^{2} \\
\dot{d}\left(t_{\tau}\right)-b_{1}-2 \times b_{2} \cdot S_{\tau0} \\
\ddot{d}\left(t_{\tau}\right)-2 \times b_{2}
\end{array}\right]
\end{array}
\end{equation}
$\mathbf{A}_1, \mathbf{B}_1$ and $\mathbf{A}_2$, $\mathbf{B}_2$ are time-dependent coefficient matrices derived from endpoint constraints \cite{29} and $S_{\tau0}$ equals $s(t_{\tau})-s(t_{0})$. This guarantees dynamic feasibility and smooth terminal constraints.

However, while quintic polynomials ensure local feasibility, discontinuities may still occur when connecting multiple segments. A composite terminal constraint $Z(\xi(\tau))$ is introduced, which incorporates kinematic continuity conditions and a perception-weighted spacing constraint within the terminal manifold of the trajectory cluster.
\begin{equation}
Z(\boldsymbol{\xi}(t_\tau)) =
\begin{bmatrix}
\dot{s}_i(t_\tau) - \dot{s}_{\mathrm{center}}(t_\tau) \\
\ddot{s}_i(t_\tau) - \ddot{s}_{\mathrm{center}}(t_\tau) \\
\dot{d}_i(t_\tau) - \dot{d}_{\mathrm{center}}(t_\tau) \\
\ddot{d}_i(t_\tau) - \ddot{d}_{\mathrm{center}}(t_\tau) \\
\alpha_p \Delta_0 - \|\boldsymbol{\xi}_i(t_\tau) - \boldsymbol{\xi}_{i-1}(t_\tau)\|
\end{bmatrix}
= \boldsymbol{0}.
\label{eq:terminal_constraint}
\end{equation}

Here, $\Delta_0$ denotes the nominal terminal spacing interval that defines the baseline sampling distance between adjacent quintic segments,  
and $\alpha_p \in [0.8, 1.2]$ is a perception-weighting coefficient reflecting local environmental complexity,  
which scales the spacing threshold according to obstacle density or road curvature. Given the states $\mathbf{\xi}(t) = \left[ s(t), \dot{s}(t), \ddot{s}(t), d(t), \dot{d}(t), \ddot{d}(t) \right]$, This formulation ensures dynamically smooth transitions and maintains spatial sparsity across the trajectory cluster. 

In summary, the HTSC Generation module outputs a compact, feasible, and smooth trajectory cluster and it establishes a reliable foundation for MTO that follows. 

\subsubsection{MTO for Visually Impaired Scenarios}
Although heuristic trajectories are dynamically feasible, they may still violate physical plausibility or induce uncomfortable motion for visually impaired users. To address this, an MTO layer imposes momentum-based constraints, aligning trajectories with human-centered requirements such as stability and smoothness during acceleration or turning maneuvers.

Building upon the system modeling and problem formulation presented in Section~\ref{p1} and Section~\ref{h11}, this study further investigates trajectory optimization in dynamic environments, specifically targeting assistive needs for visually impaired individuals. The continuous time state space model of the system and the linear observation structure of the output function can be expressed as \cite{32}:
\begin{equation}
\begin{aligned}
\dot{\boldsymbol{\xi}}(t) & =\left[\begin{array}{ll}
A & \mathbf{0} \\
\mathbf{0} & A
\end{array}\right] \boldsymbol{\xi}(t)+\left[\begin{array}{ll}
B & \mathbf{0} \\
\mathbf{0} & B
\end{array}\right] \boldsymbol{u}(t) \\
\boldsymbol{y}(t) & =\left[\begin{array}{ll}
C & \mathbf{0} \\
\mathbf{0} & C
\end{array}\right] \boldsymbol{\xi}(t)+\left[\begin{array}{ll}
D & \mathbf{0} \\
\mathbf{0} & D
\end{array}\right] \boldsymbol{u}(t) \in \mathcal{R}(t) \\
& =E \boldsymbol{\xi}(t)
\end{aligned}
\end{equation}
where
\begin{equation}
A=\left[\begin{array}{lll}
0 & 1 & 0 \\
0 & 0 & 1 \\
0 & 0 & 0
\end{array}\right], B=\left[\begin{array}{l}
0 \\
0 \\
1
\end{array}\right]
\end{equation}

The objective of trajectory optimization is framed as finding an optimal trade-off between safety and efficiency. This can be formalized as:
\begin{equation}
S = \lim_{N \to \infty} \frac{1}{N} \sum_{k=0}^{N-1} J(\boldsymbol{u}(t_k), \boldsymbol{\xi}(t_k), \mathcal{T})
\end{equation}

On this basis, we introduce a cost modeling framework grounded in the Lagrangian function $L_i$, together with a dynamic evaluation function $S_{\text{uncertainty}}$ that reflects the system’s ability to maintain stability and regulate uncertainty within a given horizon:
\begin{equation}
S_{\text{uncertainty}} = (t_\tau - t_0) \cdot \mathbb{E}_{t \in [t_0, t_\tau]} \left[ L_i(t) \right]
\end{equation}
where
\begin{equation}
L_i(t) =  E_{\text{motion},i}(t)-E_{\text{guidance},i}(t)
\end{equation}

In this formulation, $E_{\text{motion},i}$ represents the kinetic energy of agent, while $E_{\text{guidance},i}$ reflects potential energy induced by environmental constraints such as road signs, speed limits, and the behaviors of surrounding agents in assistive navigation scenarios. This formulation allows agent dynamics to be expressed in terms of energy trade-offs and provides a basis for optimal control:
\begin{equation}
\begin{aligned}
\boldsymbol{u}^* = \arg \min_{\boldsymbol{u}_{0:N}} \Bigg[ & \int_{t_0}^{t_0 + N\Delta t} L_i\left(\boldsymbol{u}(t), \boldsymbol{\xi}(t), \mathcal{T} \right) \, dt  + Z\left(\boldsymbol{\xi}(t_0 + N\Delta t), \mathcal{T} \right) \Bigg]
\end{aligned}
\end{equation}

Considering dynamic interactions in traffic environments, the system is further modeled as a set of binary agent-interaction subsystems, primarily reflecting car-following and lane-changing with directional asymmetry. MTO is then constructed, incorporating longitudinal speed limits and lateral road boundaries, as illustrated in Fig. \ref{fig4}. The resulting Lagrangian formulation $L_{i}(t)$ captures the trade-off between motion energy, guidance interaction, social compliance, and perceptual uncertainty as follows:
\begin{figure}
\centerline{\includegraphics[width =.5\textwidth]{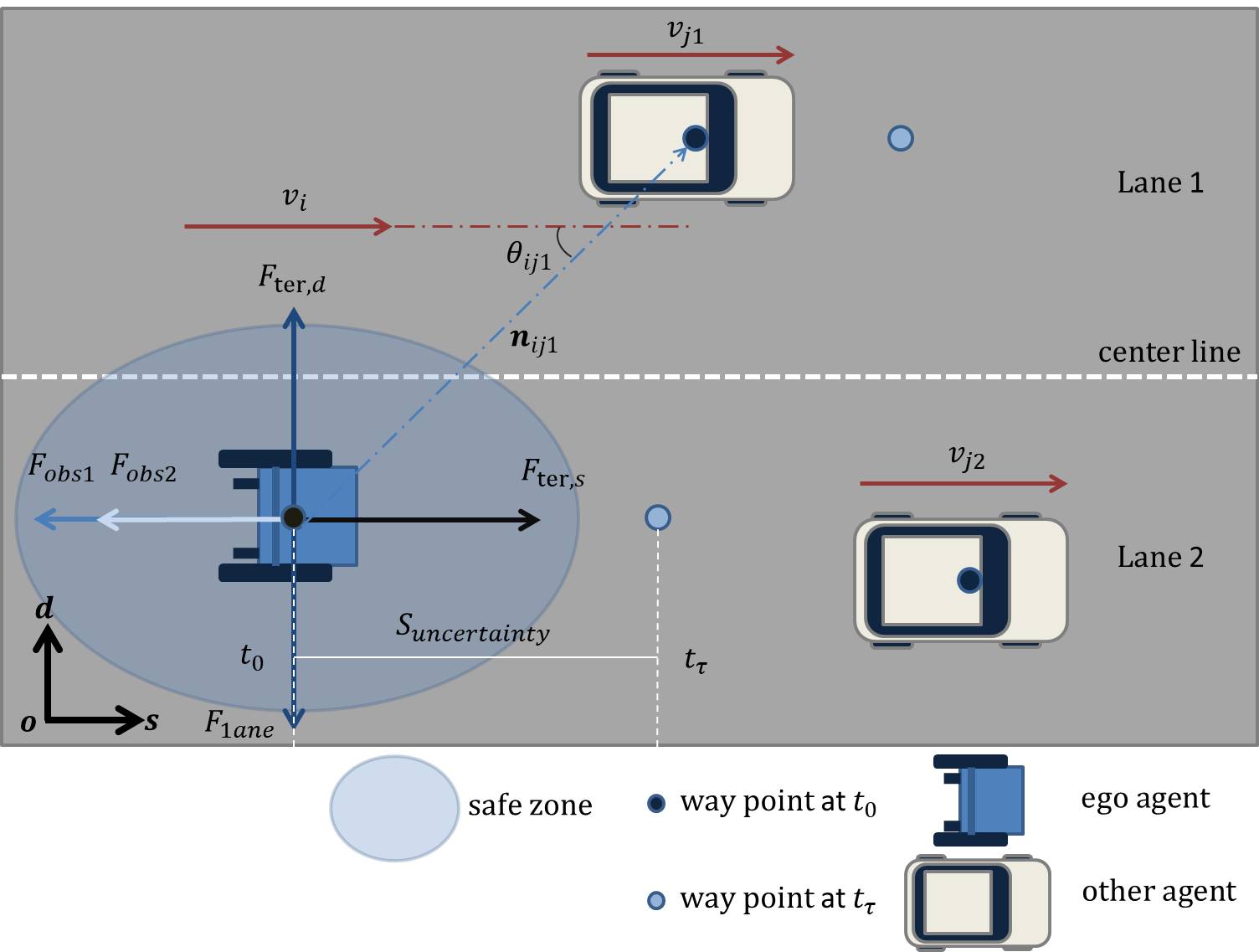}}
\caption{Schematic of MTO process for visually impaired scenario.}
\label{fig4}
\end{figure}
\begin{equation}
\begin{aligned}
L_{i}(t)= & E_{\text{motion},i}(t)-E_{\text{guidance},i}(t) \\
= & \underbrace{\frac{1}{2} m_{i}\left\|v_{i}(t)\right\|^{2}}_{\text {locomotion energy }}-\underbrace{F_{\text {asst }}(t) \cdot v_{i}(t)}_{\text {haptic/audio interaction }} \\
& -\underbrace{F_{\text {obs }}(t) \cdot\left(v_{i}(t)-v_{j}(t)\right)}_{\text {social compliance }}+\underbrace{\lambda_{s}\left\|\dot{v}_{i}(t)\right\|^{2}}_{\text {smoothness regularization }} \\
& +\underbrace{\lambda_{u} \cdot \operatorname{Tr}\left[\Sigma_{i}^{\text {perception }}(t)\right]}_{\text {perceptual uncertainty penalty }}
\end{aligned}
\label{eq19}
\end{equation}

Where $m_i$ is the mass of vehicle $i$, while $v_i(t)$ and $v_j(t)$ denote the velocities of agent $i$ and its neighboring agent $j$, respectively. ${F}_{\text{asst}}(t)$ and ${F}_{\text{obs}}(t)$ represent external forces arising from guidance systems and social interactions. The last two terms model regularization effects, where $\lambda_s$ penalizes abrupt accelerations to ensure smoothness, while $\lambda_u$ regulates perceptual uncertainty via the trace of the perception covariance matrix $\Sigma_i^{\text{perception}}(t)$.

Based on perceptual field theory \cite{33,34}, social navigation is further modeled by an anisotropic interaction force:
\begin{equation}
\boldsymbol{F}_{\text{obs}}^{i}(t)
= \sum_{j \in \mathcal{N}_i}
\frac{1}{2}\, m_{j}\,\|v_{j}(t)\|^{2}\,
r_{ij}(t)
\left(\frac{1}{r_{s}^{2}(t)}-\frac{1}{r_{d}^{2}(t)}\right)
\mathbf{n}_{ij}(t)
\label{eq21}
\end{equation}

where
\begin{equation}
\cos\theta_{ij}(t)
= \frac{v_{i,s}(t)}{|v_{i,s}(t)|+\varepsilon}\, \mathbf{n}_{ij,s}(t),
\qquad (0\le\theta_{ij}\le\pi),
\label{eq:theta_def}
\end{equation}

\begin{equation}
\begin{aligned}
r_s(t) &= r_{ij,s}(t) + B_{\!safe,s}\bigl[\,1+\eta_s\,(1-\cos\theta_{ij}(t))\bigr], \\
r_d(t) &= r_{ij,d}(t) + B_{\!safe,d}\bigl[\,1+\eta_d\,\sin^2\theta_{ij}(t)\bigr].
\end{aligned}
\label{eq:rs_rd_def}
\end{equation}

The longitudinal and lateral thresholds are angle-modulated by the relative viewing angle $\theta_{ij}$, 
where $(1-\cos\theta_{ij})$ and $\sin^2\theta_{ij}$ weights enhance forward sensitivity and attenuate off-axis responses. The anisotropy factors $\eta_s$ and $\eta_d$ control the strength of directional modulation. Eq.~(\ref{eq21}) yields a directionally modulated collision-avoidance behavior that captures forward-focused awareness and risk asymmetry, enabling deceleration-first avoidance consistent with visually impaired mobility.

The guidance-related interaction term $\boldsymbol{F}_{\text {asst }}(t)$ comprises two components: the terrain-adaptive driving force $F_{\mathrm{ter}}$ derived from slope-aware system commands, and the lateral lane-constraining force $F_{\mathrm{li}}(t)$ that models tactile or auditory lane boundary feedback \cite{35}. 
\newcommand{\sgneps}[1]{\frac{#1}{\sqrt{#1^2+\varepsilon^2}}} % small epsilon > 0
\begin{equation}
\boldsymbol{F}_{\text {asst}}(t)
= \underbrace{m_{i} g \sin \varphi_{i}(t)}_{\text{terrain-adaptive force}} \cdot \mathbf{e}_{s}
+ \underbrace{m_{i} B_{\text{type}}\!\left(\tfrac{w}{2}-|d|\right)\sgneps{d}}_{\text{lateral lane feedback}} \cdot \mathbf{e}_{d}
\end{equation}

Specifically,
\begin{equation}
\boldsymbol{F}_{\text{asst}}(t) = 
\begin{bmatrix}
F_{\text{assist},s}(t) \\
F_{\text{assist},d}(t)
\end{bmatrix}
=
\begin{bmatrix}
- F_{\mathrm{ter} ,s}(t) \\
F_{\mathrm{lane}}(t) - F_{\mathrm{ter},d}(t)
\end{bmatrix}
\end{equation}
where $\sin \varphi_{i}(t)=\lambda B_{bump}\frac{v_{desire}(t)}{v_{law}(t)}$, with $v_{desire}(t)$ and $v_{law}(t)$ denoting the desired velocity and the prescribed velocity, respectively, while $B_{bump}$ can mitigate the impact of bumps. $B_{type}$ denotes the line shape of the road boundary, and the value of this parameter is fixed due to dataset characteristics \cite{36,37}. $w$ is the lane width and $\varepsilon$ denotes a small positive constant that ensures smoothness and numerical stability around $d=0$. The unit vectors $\boldsymbol{e}_{s}$ and $\boldsymbol{e}_{\boldsymbol{d}}$ represent the longitudinal and lateral axes in the local motion frame.

Finally, the Lagrangian functions of longitudinal $L_s$ and lateral $L_d$ motion are decomposed and optimized within the Frenet coordinate framework through a backtracking approach, resulting in a spatiotemporally coupled trajectory planner.
\begin{equation}
\left\{
\begin{aligned}
L_s &= \int_{t_0}^{t_\tau} \bigg[
\frac{1}{2} m_i v_{i,s}^2(t)
- \left( F_{\mathrm{ter} ,s}(t) + \boldsymbol{F}_{\text{obs},s}^{i}(t) \right) v_{i,s}(t) \\
&\quad + \lambda_s \dot{v}_{i,s}^2(t)
+ \lambda_u \cdot \text{Tr}\left[\Sigma_i^{\text{perception}}(t)\right]
\bigg] \mathrm{d}t \\
L_d &= \int_{t_0}^{t_\tau} \bigg[
\frac{1}{2} m_i v_{i,d}^2(t)
- \left( F_{\mathrm{lane}}(t) - F_{\mathrm{ter} ,d}(t) + \boldsymbol{F}_{\text{obs},d}^{i}(t) \right) \\
& \quad\cdot v_{i,d}(t)+ \lambda_s \dot{v}_{i,d}^2(t)
+ \lambda_u \cdot \text{Tr}\left[\Sigma_i^{\text{perception}}(t)\right]
\bigg] \mathrm{d}t
\end{aligned}
\right.
\end{equation}

The Euler-Lagrange equations are utilized for trajectory solution:
\begin{equation}
\frac{\mathrm{d}}{\mathrm{~d} t}\left(\frac{\partial L}{\partial \dot{\boldsymbol{\xi_i}}}\right)-\left(\frac{\partial L}{\partial \boldsymbol{\xi_i}}\right)+\frac{\mathrm{d}^{2}}{\mathrm{~d} t^{2}}\left(\frac{\partial L}{\partial \ddot{\boldsymbol{\xi_{i}}}}\right)=0
\end{equation}
where the generalized coordinate $\boldsymbol{\xi_i}(t)$ encapsulates both longitudinal and lateral kinematics in the Frenet frame. This formulation enables the trajectory planner to reason jointly about directional feasibility and optimality. The above formulation yields a trajectory planner capable of jointly reasoning about directional feasibility and optimality. 
\begin{remark}
Unlike conventional optimization, the framework introduces MTO as a core scheme to prevent oscillations and ensure stable maneuvering. On this basis, several human-centered innovations are incorporated into the Lagrangian model. The guidance interaction term $\boldsymbol{F}{\text{guidance}}(t)$ embeds slope-sensitive modulation $B{\text{bump}}$, mitigating instability on uneven terrains. The lateral correction force encodes infrastructure-aware modulation $B_{\text{type}}$, improving adaptability to diverse road geometries. The crowd repulsion term introduces an informed asymmetric safety buffer, enlarging protective zones to reflect the limited situational awareness of visually impaired users. Finally, the perceptual uncertainty penalty $\text{Tr}[\Sigma_i^{\text{perception}}(t)]$, balanced by $\lambda_s$ and $\lambda_u$, enables robustness against sensor noise while preserving comfort-oriented smoothness. Collectively, MTO establishes a unified human-centric optimization mechanism, embedding stability, adaptability, social safety, and robustness into the trajectory planning process.
\end{remark}

\subsubsection{Trajectory Evaluation for Visually Impaired Scenarios}
Building on the stability guarantees established by MTO, the stage of DCMM performs trajectory evaluation in the Cartesian frame. Beyond kinematic and collision checks \cite{38}, this stage integrates a semantically structured cost model to transform raw trajectory candidates into compact, human-aligned representations. Formally, each trajectory $x(t)$ is evaluated as:
\begin{equation}
J_s = \sum_{i=1}^{n} \lambda_i^{\text{DRL}} \cdot \phi_i(x(t))
\end{equation}
where, $\phi_i(x(t))$ denotes key evaluation metrics, and $\lambda_i^{\text{DRL}}$ represents dynamic weighting coefficients predicted by the policy network based on context-aware features and user preferences. The instantiated terms include Acceleration Energy (AE), Jerk Minimization (JM), Velocity Optimality (VO), Obstacle Proximity Penalty (OPP), Risk Field Penalty (RFP), and Social Compliance (SC). 
\begin{equation}
\left\{
\begin{array}{ll}
\phi_1(x) = \int a_x^2(t) \, dt & \text{(AE)} \\
\phi_2(x) = \int j^2(t) \, dt & \text{(JM)} \\
\phi_3(x) = \int \left| v_x(t) - v_{\text{desire}}(t) \right| \, dt + & \text{(VO)} \\
\quad \left( v_x(t_f) - v_{\text{desire}}(t_f) \right)^2 & \\
\phi_4(x) = \int \frac{1}{\Delta x_{\text{obs}}^2(t)} \, dt & \text{(OPP)} \\
\phi_5(x) = \int f(x, 0, \Sigma_{\text{rot}}) \, dx \, dt & \text{(RFP)} \\
\phi_6(x) = \int \frac{1 - t/T}{d_{\mathrm{M}}(v_j(t), v_x(t), \Sigma(t))} \, dt & \text{(SC)}
\end{array}
\right.
\end{equation}

\begin{remark}
    Unlike conventional feasibility filters that rely on sparse terminal rewards \cite{41}, this stage of DCMM embeds semantically interpretable cost terms into the evaluation process. By transforming feasibility checks into dense, human-centered learning signals, it enables the downstream DRL module to achieve stability, adaptability, interpretability, and safety—properties that are essential for visually impaired scenarios.

\end{remark}

\subsection{Design of Residual-enhanced DRL Framework}
\begin{figure}
\centerline{\includegraphics[width =.5\textwidth]{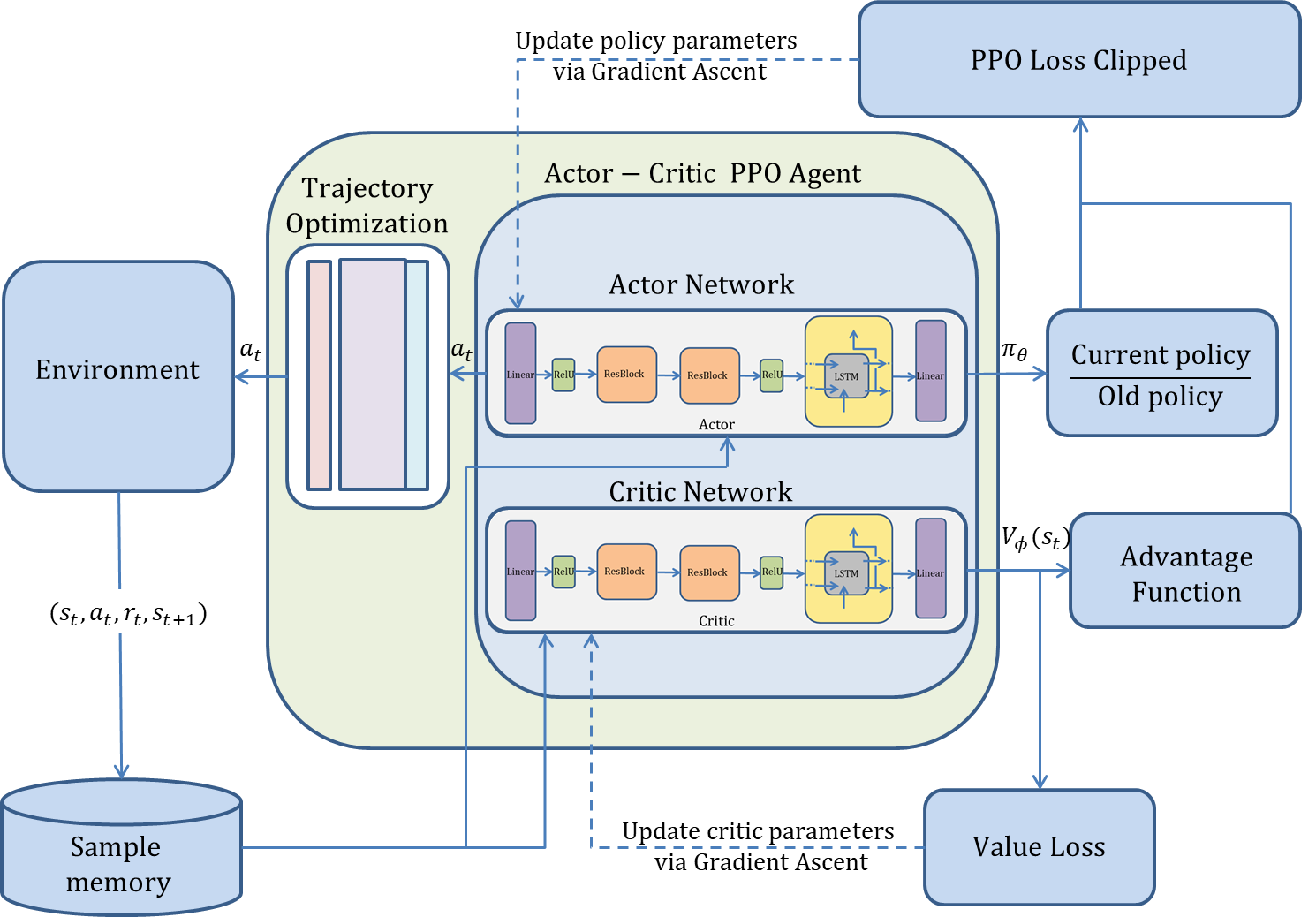}}
\caption{Proposed Residual-enhanced DRL framework with temporal modeling based on PPO.}
\label{fig5}
\end{figure}

While HTSCMOE ensures feasibility, effective deployment in assistive scenarios also requires learning policies that are expressive, context-aware, and robust under uncertainty. As the back-end of the framework, the DRL extends PPO \cite{43} by integrating residual temporal modeling, context-aware state–action abstraction, and hierarchical reward shaping. As shown in Figure \ref{fig5}, this design overcomes restricted representational capacity, opaque decision rules, and sparse rewards, thereby aligning policy learning with the safety, comfort, and interpretability demands of visually impaired navigation.

\subsubsection{Residual-enhanced and Temporal Feature Modeling for Policy and Reward Networks}
Since shallow MLPs fail to capture rich semantics and long-term temporal dependencies, the policy and reward networks are designed with dual-stacked residual blocks(ResBlocks) \cite{49} and a lightweight LSTM unit \cite{46}. Residual shortcuts improve gradient flow and stability, while the LSTM captures sequential patterns in latent features. This combination enables robust long-horizon decision-making without excessive computational cost. 

While two residual blocks strengthen static feature learning, temporal consistency is further captured by inserting a lightweight LSTM after residual layers:
\begin{equation}
h_t = \operatorname{LSTM}(x_t^{\text{Res}}, h_{t-1}, c_{t-1})
\end{equation}
where $x_t^{\text{Res}}$ is the output from stacked residual layers. This design choice enables the LSTM to operate on high-level encoded features rather than raw inputs, improving efficiency and interpretability.

\begin{figure}
\centerline{\includegraphics[width =.5\textwidth]{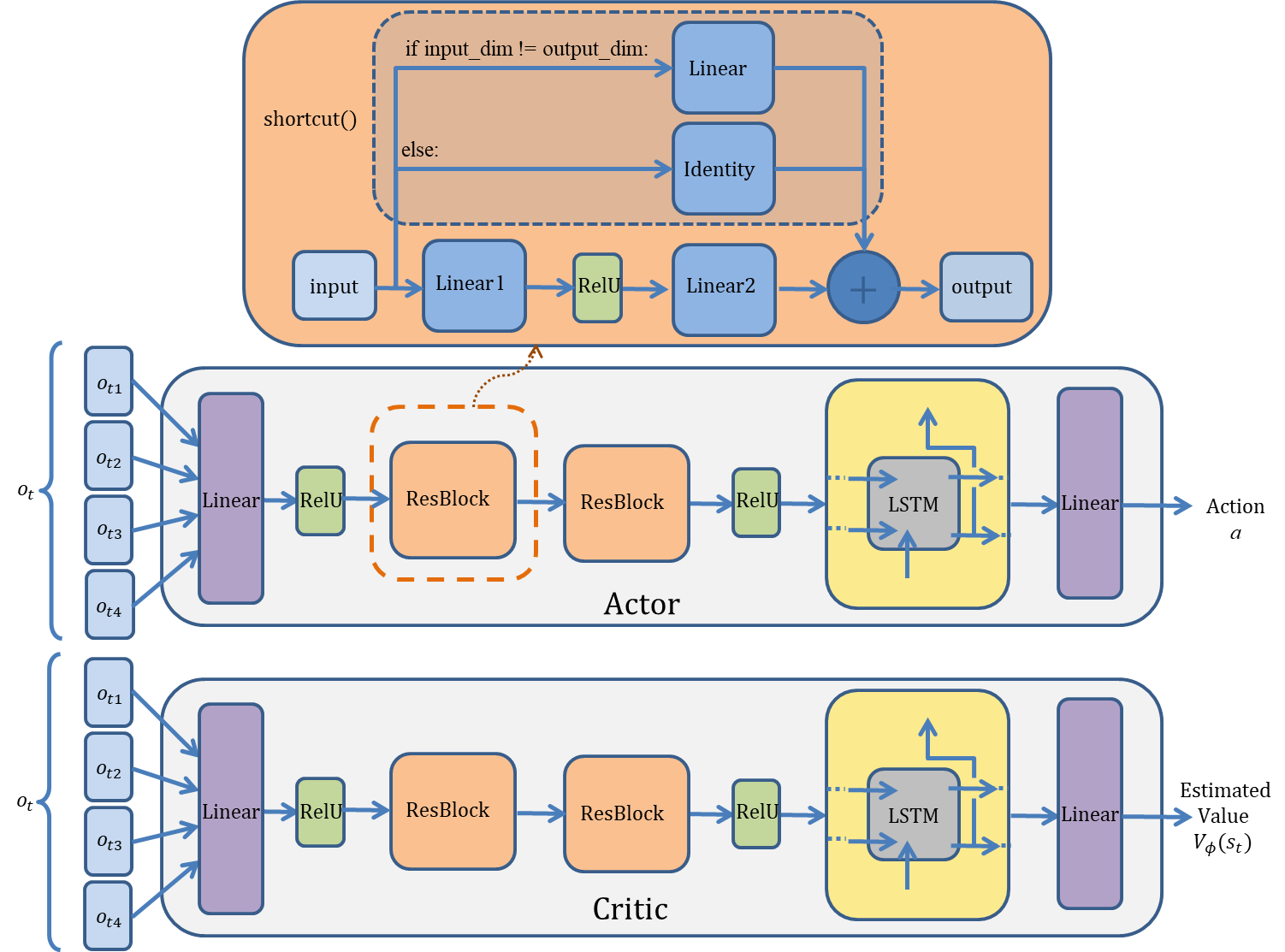}}
\caption{Residual-enhanced temporal architecture for Actor Network and Critic Network, Where the policy and reward networks correspond to the actor and critic components, respectively.}
\label{fig6}
\end{figure}

The LSTM module is designed with controlled dimensionality, such as using eight hidden units, to facilitate deployment on resource-constrained platforms. The final outputs, whether action logits in the policy network or scalar reward estimates in the reward network, are generated through simple linear projection layers.
\begin{equation}
\text{output} = \operatorname{Linear}(h_t)
\end{equation}
\begin{remark}
    The policy and value branches in the proposed architecture employ two consecutive residual blocks tailored for fully connected networks, where identity or linear projections are used as shortcut connections to ensure dimensional consistency with minimal parameter overhead. By avoiding heavy normalization and excessive depth, this design preserves the advantages of residual learning, including improved gradient flow, stability, and feature expressiveness, without unnecessary complexity, thus facilitating faster convergence and better generalization under limited training budgets. On top of this residual backbone, a compact LSTM module with only 8 hidden units is appended to capture temporal dependencies in agent–environment interactions. Operating on high-level encoded features rather than raw inputs, the LSTM models sequential patterns while maintaining low inference latency, balancing expressiveness with efficiency. This residual–temporal integration is particularly suited to real-time assistive navigation, where stability, responsiveness, and deployment feasibility are critical requirements.
\end{remark}
While both networks share the same structural design, their parameters are fully independent, ensuring functional decoupling between decision-making and reward modeling for robust planning in visually impaired scenarios. The overall network designs are illustrated in Figure \ref{fig6}, and the architectures are shown in Table \ref{table2} and \ref{table3}.

\begin{table*}[htbp]
\centering
\caption{Policy Network Architecture}
\begin{tabular}{llll}
\toprule
\textbf{Module} & \textbf{Sub-components} & \textbf{Output Dimensions} & \textbf{Activation Function} \\
\midrule
Feature Extractor & FlattenExtractor & - & - \\
Policy MLP & Linear (4$\rightarrow$32), ResBlock ($\times$2), ReLU & 100 & ReLU \\
Policy Memory Unit & LSTM (100$\rightarrow$8) & 8 & - \\
Policy Output Head & Linear (8$\rightarrow$2) & 2 & - \\
\bottomrule
\end{tabular}
\label{table2}
\end{table*}

\begin{table*}[htbp]
\centering
\caption{Reward Network Architecture}
\begin{tabular}{llll}
\toprule
\textbf{Module} & \textbf{Sub-components} & \textbf{Output Dimensions} & \textbf{Activation Function} \\
\midrule
Feature Extractor & FlattenExtractor & - & - \\
Value MLP & Linear (4$\rightarrow$32), ResBlock ($\times$2), ReLU & 100 & ReLU \\
Value Memory Unit & LSTM (100$\rightarrow$8) & 8 & - \\
Value Output Head & Linear (8$\rightarrow$1) & 1 & - \\
\bottomrule
\end{tabular}
\label{table3}
\end{table*}

\subsubsection{Context-Aware Policy Modeling for Visually Impaired Scenarios}
Building on the first-stage evaluation in DCMM, the context-aware policy modeling framework refines candidate trajectories through DRL-driven adaptation. It bridges perceptual limitations with human-centric decision objectives, and realizes the second stage of DCMM via three tightly coupled components, as follows.

1) \textbf{Observation Space with Semantic and Risk Encoding}. 
The observation space is constructed around cognition-based environmental understanding \cite{51}, thereby overcoming the limitation of absent direct vision. By integrating motion estimates, obstacle proximity, lane semantics, and trajectory feasibility descriptors into the agent’s state vector $o_t^{\text{cog}}$, the policy gains semantically structured cues that enhance interpretability and improve decision robustness under perceptual constraints.
\begin{equation}
\begin{aligned}
o_{t}^{\operatorname{cog}}=\{ & \hat{v}_{t}, \hat{a}_{t}, \hat{\omega}_{t}, \theta_{t}, \\
& d_{\text {goal }}, t_{\text {remain }}, \mathbf{1}_{\text {goal }}, \\
& \phi_{\text {lane }}^{L}, \phi_{\text {lane }}^{R}, \rho_{\text {obs }}, \\
& \left.\eta_{\text {valid }}, F_{\text {feas }}, \mu_{F}, \sigma_{F}, p_{\text {risk }}^{\operatorname{cog}}\right\}
\end{aligned}
\end{equation}
where, variables such as $\hat{v}_{t}$ , $\rho_{\text {obs }}$ , and $p_{\text {risk }}^{\operatorname{cog}}$ represent estimated motion states, obstacle proximity, and perceived collision risk, respectively. Features like $F_{\text {feas}}$ , $ \mu_{F}$, and $\sigma_{F}$ quantify the feasibility and variability of candidate trajectories, thereby emulating how visually impaired users cognitively assess navigability. Semantic lane features $\phi_{\text {lane }}^{L / R}$ substitute direct lane-line detection with probabilistic alignment cues. This structured observation design enables the policy to simulate human-like navigation under constrained perception.
\begin{remark}
    The proposed cognitive observation space $o_t^{\text{cog}}$ not only provides semantically structured and compact representations, but also establishes the input foundation for DCMM. Environmental cues are refined, such as decomposing road context into left and right lane directions, while trajectory-related cost and risk features are integrated into a unified feasibility-aware representation. This design reduces observation dimensionality and enhances learning efficiency, better aligning with the structured, focused perception style of visually impaired users.
\end{remark}
2) \textbf{Action Space via Cost Weight Adaptation}. Rather than outputting raw control signals, the policy directly modulates the dynamic cost weights of DCMM $\lambda_i^{\text{DRL}}$. This formulation transforms opaque action choices into interpretable adjustments of the underlying cost model, enabling the agent to adaptively trade off safety, comfort, and efficiency \cite{84} in real time. The update at step $t$ is defined as:
\begin{equation}
\lambda_i^{t} = \text{clip}\left(\lambda_i^{t-1} + \Delta \lambda_i^{\text{DRL}}, \lambda_i^{\min}, \lambda_i^{\max}\right)
\end{equation}
By regulating weights instead of low-level motor commands, the policy achieves context-aware adaptation \cite{85}, such as slowing down near dense obstacles, prioritizing comfort in complex geometries, or accelerating in open and safe regions. The clipped update scheme ensures both interpretability and temporal stability, which are crucial for trustworthy assistive navigation in real-world visually impaired scenarios.

3) \textbf{Hierarchical Reward Design for Multi-Level Guidance}. A hierarchical reward design provides dense, semantically structured feedback to guide the second-stage cost modeling of DCMM. Organized into task completion, behavioral quality, and risk sensitivity layers, this reward structure ensures that policy learning remains aligned with human-centric navigation priorities. The overall reward at each timestep $r_t$ is computed as:
\begin{equation}
r_{t}=\sum_{l \in\{\text { task }, \text { behav, risk }\}} \lambda_{l} \cdot \sum_{k} r_{t}^{(l, k)}
\end{equation}
where $\lambda_{l}$ indexes the semantic layer corresponding to task completion \textit{task}, behavioral quality \textit{behav}, and risk sensitivity \textit{risk}, and denotes the configurable layer-wise weight that reflects user-specific preferences or environmental context.

At the task level, rewards ensure timely and feasible arrival while penalizing collisions or timeouts, reflecting the primary concern of safety for visually impaired users. Behavioral rewards quantify smoothness, progress efficiency, and deviation from reference paths, thereby capturing comfort and interpretability \cite{86}. Risk-sensitive rewards penalize unsafe actions such as over-speeding or proximity to obstacles, mirroring heightened safety awareness in assistive scenarios. The structured reward mechanism is summarized in Table \ref{table4}.

\begin{remark}
    Unlike conventional reward designs that rely on sparse outcomes or ad hoc shaping, the proposed hierarchical scheme embeds DCMM’s semantic priorities directly into policy learning. By unifying task success, behavioral quality, and risk minimization, it provides interpretable dense signals that accelerate convergence, enable flexible customization, and ensure safety- and comfort-aware decision-making for visually impaired navigation.
\end{remark}
% 需要的宏包：
% \usepackage{array,multirow,booktabs}

\begin{table*}[htbp]
\renewcommand{\arraystretch}{1.12}     % 统一行距
\setlength{\extrarowheight}{0.6pt}     % 轻微抬高改善视觉
\setlength{\tabcolsep}{4pt}            % 列间距
\centering
\caption{Context-Aligned Hierarchical Reward Design for Visually Impaired Navigation}
\begin{tabular}{
  >{\raggedright\arraybackslash}m{3.4cm}  % Reward Layer
  >{\raggedright\arraybackslash}m{2.2cm}  % Objective
  >{\raggedright\arraybackslash}m{2.9cm}  % Reward Signal
  >{\centering\arraybackslash}m{2.6cm}    % Mathematical Expression
  >{\raggedright\arraybackslash}m{4.6cm}  % Cognitive Description
}
\toprule
\textbf{Reward Layer} & \textbf{Objective} & \textbf{Reward Signal} & \makecell{\textbf{Mathematical}\\ \textbf{Expression}} & \textbf{Cognitive Description} \\
\midrule

\multirow{2}*[-0.2ex]{\textbf{Task Guidance Layer}} % 轻微下移, 让两行居中更自然
& \multirow{2}{=}{Task completion, path feasibility}
& Task result feedback
& \( r_t^{\text{comp}} \)
& \shortstack[l]{Success, timeout, or infeasibility\\judgment from global cognition.} \\

& 
& Completion timing feedback
& \( r_{\text{goal}}^{+},\, r_{\text{goal}}^{-} \)
& \shortstack[l]{Reward early task completion;\\penalize delay in cognition-based\\planning.} \\

\multirow{4}*[-0.4ex]{\textbf{Behavior Guidance Layer}}
& \multirow{4}{=}{Driving quality, efficiency, comfort}
& Reference alignment deviation
& \( \delta_{\text{align}} \)
& \shortstack[l]{Penalize lateral deviation from\\the preferred navigation corridor.} \\

& 
& Speed deviation penalty
& \( \delta_v^{\text{smooth}} = |v_t - v_{\text{target}}| \)
& \shortstack[l]{Encourage speed consistency for\\smoother user-perceived motion.} \\

& 
& Goal distance reduction
& \( \delta_{g}^{\text{prog}} \)
& \shortstack[l]{Reward forward progress toward the\\perceived goal direction.} \\

& 
& Cost-optimality bias
& \( \Delta C^{\text{cog}} = C_{\text{min}} - C_{\text{ego}} \)
& \shortstack[l]{Encourage selection of cognitively\\efficient trajectories.} \\

\multirow{2}*[-0.2ex]{\textbf{Risk Avoidance Layer}}
& \multirow{2}{=}{Robustness, risk minimization}
& Self-action risk
& \( \rho_{\text{self}}^{\text{risk}} \)
& \shortstack[l]{Penalize over-speeding, unsafe\\acceleration, or collision probability.} \\

& 
& Environmental risk
& \( \rho_{\text{obs}}^{\text{risk}} \)
& \shortstack[l]{Penalize proximity to static obstacles\\or road edges.} \\
\bottomrule
\end{tabular}
\label{table4}
\end{table*}

\section{Results \& analysis}
\label{r}
\subsection{Environment and training setup}
We evaluate the proposed framework using the CommonRoad benchmark \cite{72}, developed by the Technical University of Munich as a standardized platform for motion planning research. While originally designed for general traffic scenario evaluation, CommonRoad provides a diverse set of structured road networks with modular topologies, dynamic agents, and temporal-spatial constraints, which makes it well suited for trajectory optimization frameworks in complex environments. 

In this work, we specifically curate low-speed urban scenarios from CommonRoad that best align with visually impaired navigation contexts. The selected subset prioritizes \textit{URBAN} and \textit{INTERSECTION} tags, includes a moderate number of dynamic obstacles ($3$–$10$ agents such as cars, cyclists, or pedestrians), and initial velocities limited to $2$–$8$ m/s. Goal specifications based on \textit{LANELET} and \textit{TIME\_STEP} ensure that the planning tasks remain consistent with real-world assistive requirements. Representative cases include left turns, oncoming traffic, and two-lane road segments, which capture key characteristics of pedestrian- and wheelchair-relevant environments while ensuring reproducibility. 

The training environment is standardized via the Gymnasium interface, with customized observation spaces and reward functions. Algorithm implementation builds on Stable-Baselines3 with PyTorch, employing a residual-enhanced Actor–Critic policy network. Training stability is monitored through reward, entropy, and value loss, with checkpointing and early-stop mechanisms ensuring efficiency.  

Experiments are conducted on a workstation with four NVIDIA RTX 4090 GPUs (24 GB each), 128 GB DDR5 memory, and 2 TB NVMe SSD. Parallel training exceeds $10^6$ steps, with the best-performing model selected after approximately $5\times 10^5$ steps. Hyperparameters are tuned for stability and efficiency, with the key settings summarized in Table \ref{table5}. 

\begin{table}[p]
\centering
\caption{Table of Hyperparameters    }
\begin{tabular}{lllc}
\hline
\textbf{Hyperparameter} & \textbf{Description} & \textbf{Value} \\
\hline
$l$& Learning rate & 0.0003 \\
$\epsilon$& Clipping hyperparameter & 0.1 \\
$\gamma$& Discount factor & 0.80 \\
$\lambda$& GAE & 0.97 \\
$B$& Batch size & 2352 \\
$N^E$& Number of training epochs & 5 \\
$c_2$& Entropy coefficient & 0.01 \\
\hline
\end{tabular}
\label{table5}
\end{table}

\subsection{Reward Evaluation}
\begin{figure*}
\centering
\begin{subfigure}{0.49\linewidth}
    \centerline{\includegraphics[width = .99\textwidth]{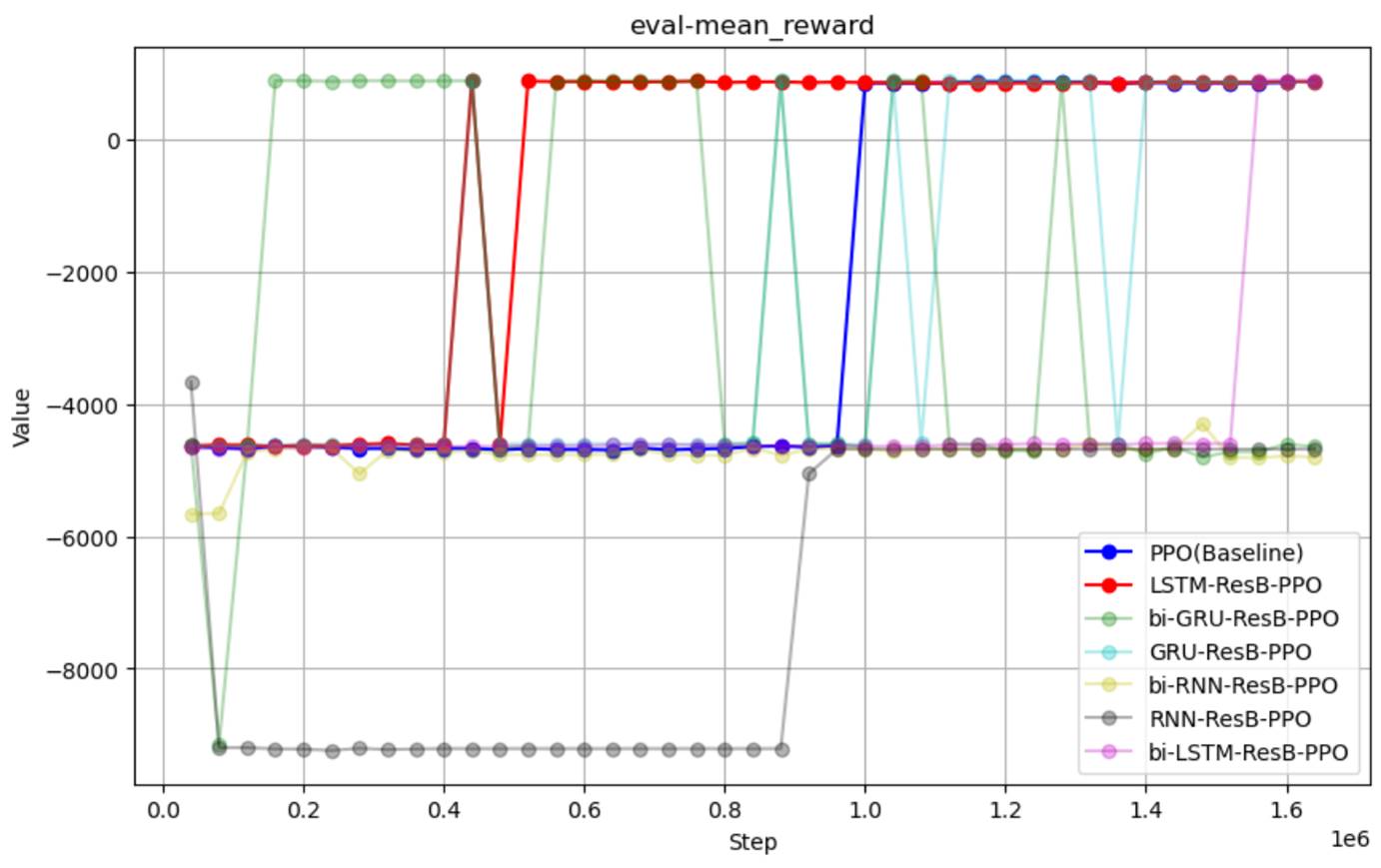}}
    \caption{}
    \label{fig7a}
\end{subfigure}
\begin{subfigure}{0.49\linewidth}
    \centerline{\includegraphics[width = .99\textwidth]{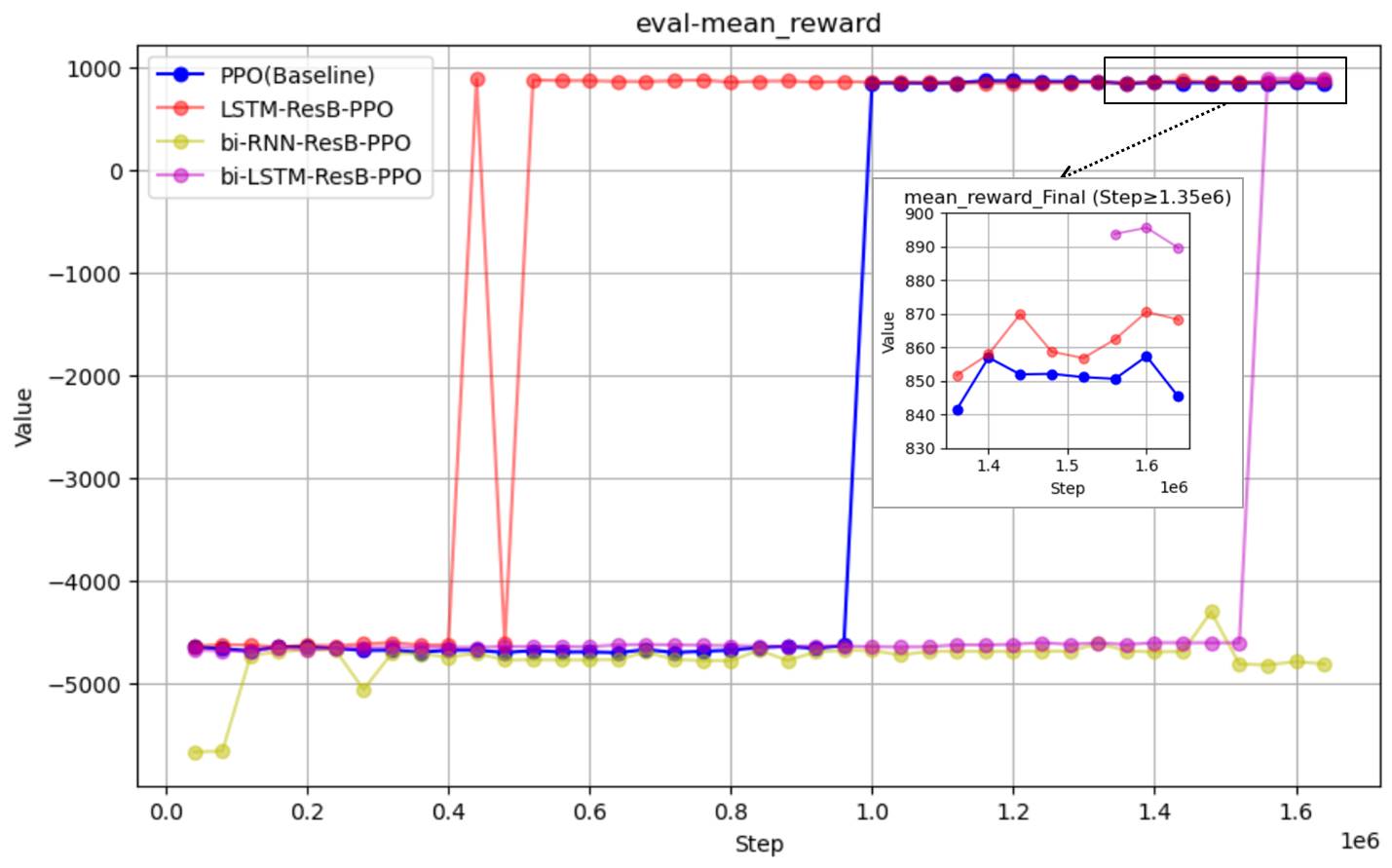}}
    \caption{}
    \label{fig7b}
\end{subfigure}
\begin{subfigure}{0.49\linewidth}
    \centerline{\includegraphics[width = .99\textwidth]{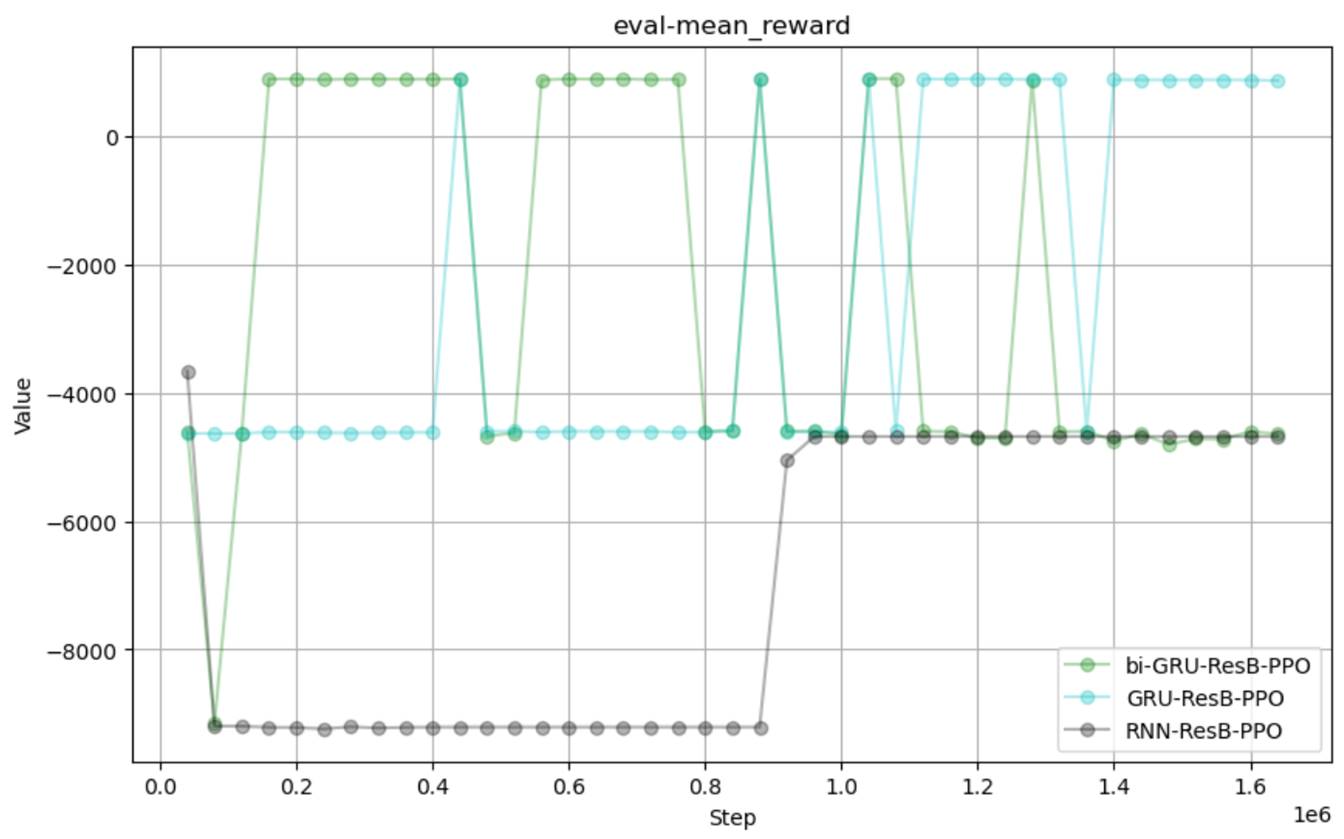}}
    \caption{}
    \label{fig7c}
\end{subfigure}
\caption{Line graph comparing the change of mean reward with the number of training steps for different algorithm training processes, where (a) shows the training performance of all algorithms, (b) shows the performance of algorithms that converged after training, and (c) shows the performance of algorithms that did not converge after training.}
\label{fig7}
\end{figure*}

Reward is a primary indicator of policy quality and learning progress. As shown in Figure \ref{fig7}, the horizontal axis represents the number of training steps and the vertical axis the mean cumulative reward. Different curves correspond to alternative temporal modeling structures. During the initial training phase, all models exhibit large fluctuations due to random exploration, indicating instability in early policy learning. As training progresses, convergence speed and stability differ significantly across models.

The proposed LSTM-ResB-PPO converges substantially faster, reaching a stable reward level around $520{,}000$ steps, whereas the baseline PPO requires approximately $1{,}000{,}000$ steps to achieve comparable stability. In terms of performance metrics, the enhanced model attains a higher final performance (FP) of $861.138$, compared to $852.301$ for the baseline, and achieves a peak reward (PR) of $877.502$ versus $869.986$. Its converged performance (CP) is likewise superior, with an average of $861.055$ against $855.022$ for the baseline. These improvements arise because the residual blocks strengthen feature representation and stabilize gradient propagation, while the LSTM module captures temporal dependencies that reduce unstable exploration, thereby accelerating convergence and improving overall reward quality.

The model also maintains robustness in terms of minimum reward (MR), remaining above $851.776$, whereas the baseline drops to $841.548$. This indicates that the enhanced architecture prevents drastic performance collapse under difficult training conditions. Regarding stability, the LSTM-ResB-PPO exhibits only a single fluctuation during training, in contrast to the GRU- and bi-GRU-based models, which suffer from five and seven major oscillations, respectively. This instability arises because such models are more sensitive to parameter tuning and fail to consistently capture long-term dependencies. Moreover, the RNN and bi-RNN variants collapse entirely into negative reward regions, with the bi-RNN model reaching a final reward as low as $-4695.173$. These results underscore that merely replacing the temporal encoder is insufficient; robust stability requires the combined effect of residual enhancement and LSTM-based modeling. 

Table \ref{table6} summarizes the comparative metrics across models, including convergence step (CS), FP, CP, PR, MR, and number of fluctuations (NF). Overall, the results confirm that the residual-enhanced LSTM structure achieves the best balance between efficiency and stability. In terms of efficiency, it converges substantially faster, as reflected by a shorter CS, and yields superior FP and PR. In terms of stability, it maintains higher MR values, exhibits fewer NF, and delivers more consistent CP. These characteristics make the proposed model particularly effective for safety-critical assistive trajectory optimization tasks, where both reliable convergence and robust policy execution are essential.

\begin{table*}[htbp]
\centering
\caption{Performance Comparison of Different Algorithms about Mean Reward}
\begin{tabular}{lcccccc}
\hline
\textbf{Algorithm} & \makecell{\textbf{CS}} & \makecell{\textbf{FP}} & \makecell{\textbf{CP}} & \textbf{PR} & \textbf{MR} & \makecell{\textbf{NF}} \\
\hline
PPO Baseline \cite{6} & 1000000 & 822.301 & 855.022 & 869.986 & 841.548 & 0 \\
LSTM-ResB PPO & 520000 & 861.138 & 861.055 & 877.502 & 851.776 & 1 \\
bi-LSTM-ResB PPO & 1560000 & -2541.737 & 893.035 & 895.653 & 889.659 & 0 \\
RNN-ResB PPO & 960000 & -4683.867 & -4683.656 & -4680.461 & -4683.867 & 0 \\
bi-RNN-ResB PPO & 1520000 & -4695.173 & -4802.599 & -4782.936 & -4815.568 & 2 \\
GRU-ResB PPO & - & 185.280 & - & - & - & 5 \\
bi-GRU-ResB PPO & - & -4687.911 & - & - & - & 7 \\
\hline
\end{tabular}
\label{table6}
\end{table*}
\subsection{Episode Length Evaluation}
\begin{figure*}
\centering
\begin{subfigure}{0.49\linewidth}
    \centerline{\includegraphics[width = .99\textwidth]{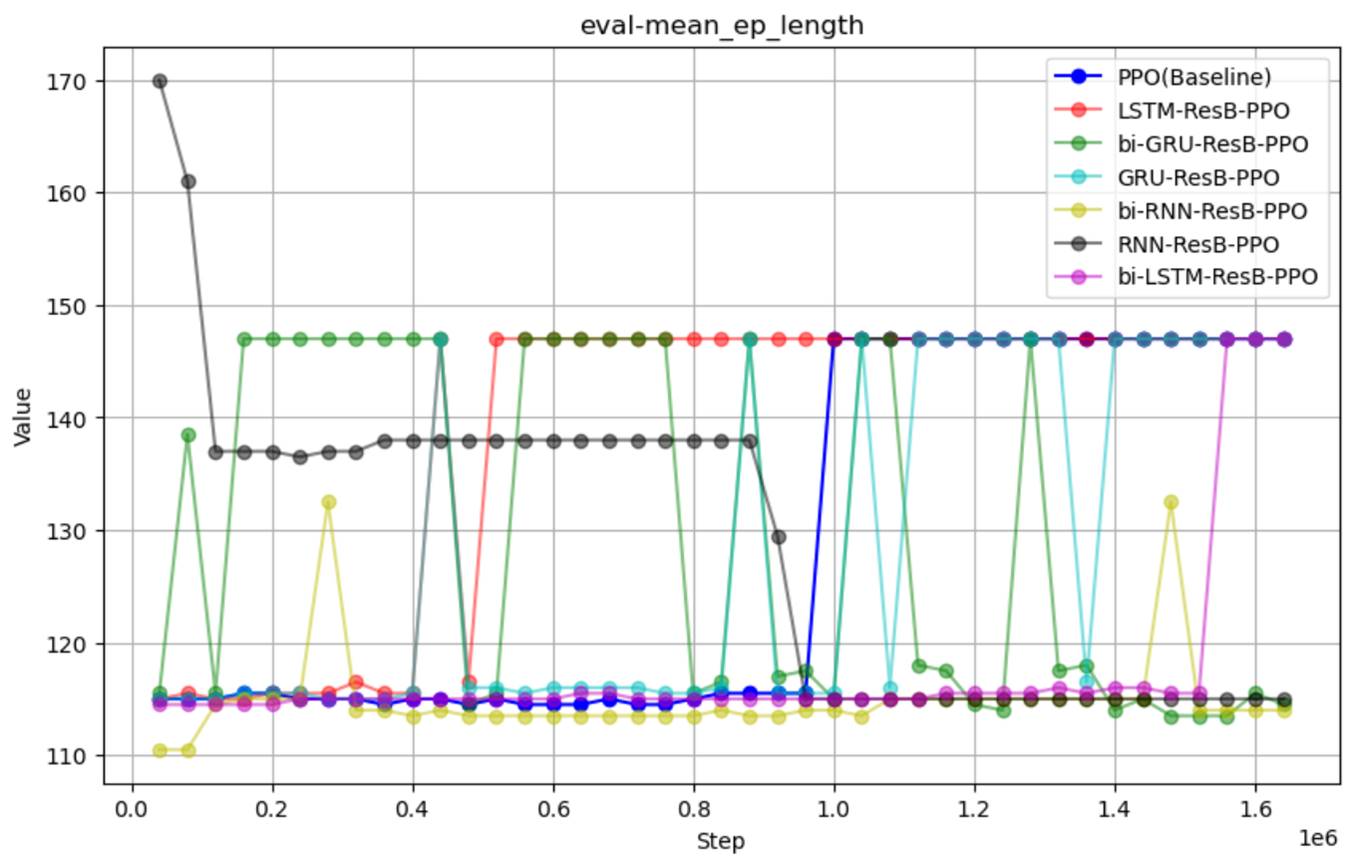}}
    \caption{}
    \label{fig8a}
\end{subfigure}
\begin{subfigure}{0.49\linewidth}
    \centerline{\includegraphics[width = .99\textwidth]{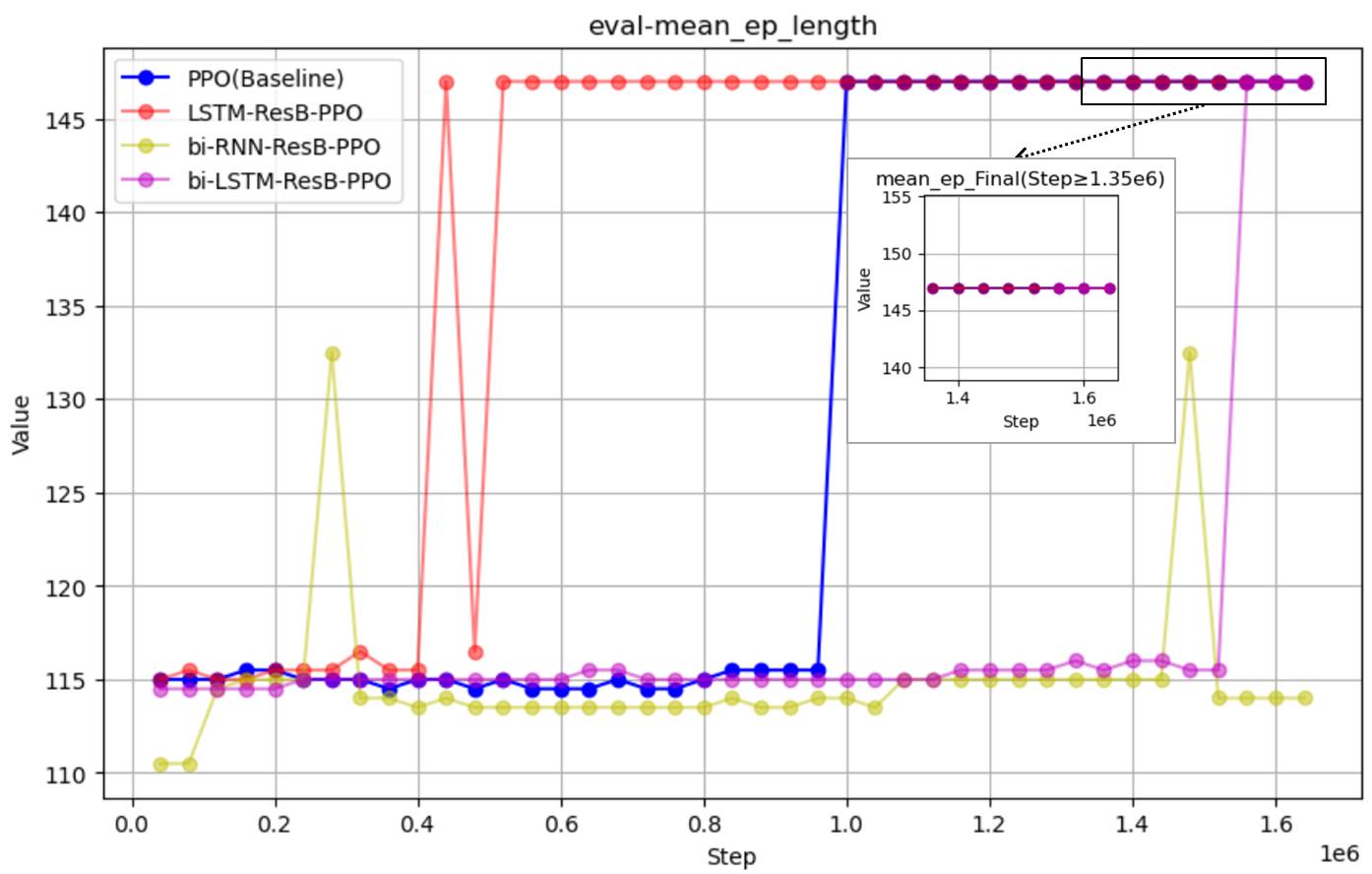}}
    \caption{}
    \label{fig8b}
\end{subfigure}
\begin{subfigure}{0.49\linewidth}
    \centerline{\includegraphics[width = .99\textwidth]{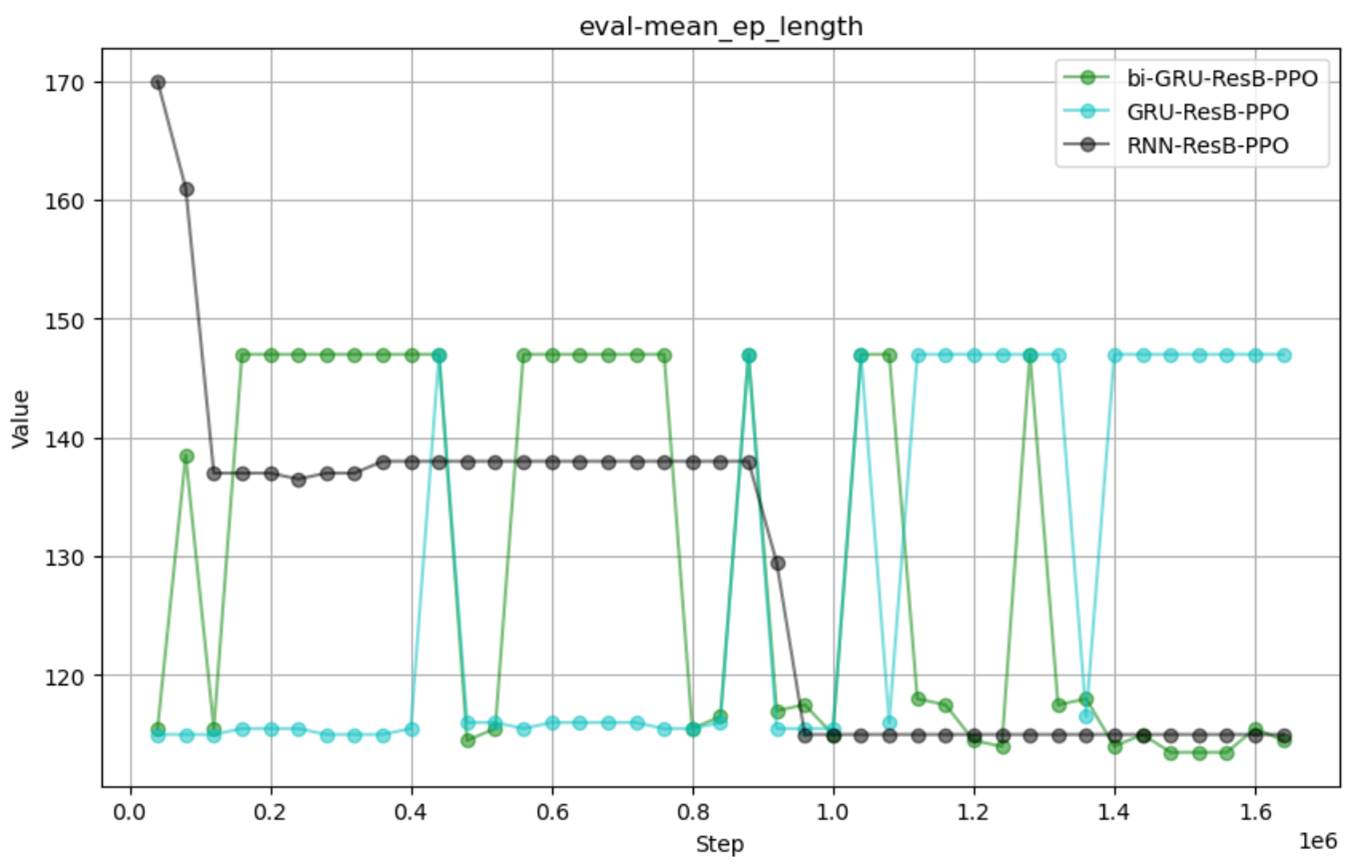}}
    \caption{}
    \label{fig8c}
\end{subfigure}
\caption{Line graph comparing the change of average ep length with the number of training steps for different algorithm training processes, where (a) shows the training performance of all algorithms, (b) shows the performance of algorithms that converged after training, and (c) shows the performance of algorithms that did not converge after training.}
\label{fig8}
\end{figure*}

Episode length reflects the efficiency of task execution, indicating how quickly the agent can reach its goal under a given policy. As shown in Figure \ref{fig8}, the horizontal axis denotes the number of training steps and the vertical axis the average episode length, with different curves representing alternative temporal encoders. During the early training phase, most models display high variance and unstable trajectories due to exploratory behavior. As training progresses, the models differ substantially in both convergence speed and execution stability.
The proposed LSTM-ResB-PPO achieves rapid convergence to a stable episode length of 147 within approximately $520{,}000$ steps, whereas the baseline PPO requires nearly $1{,}000{,}000$ steps to reach the same level. Importantly, this efficiency is achieved without sacrificing reward performance, because the residual blocks provide stable feature propagation and the LSTM module captures long-term dependencies, thereby guiding the agent toward efficient strategies earlier in the training process. 

The model also demonstrates robustness in terms of peak and minimum episode lengths. It maintains a consistent episode length of 147 at convergence, with minimal deviation across training runs, while incurring only a single fluctuation. By contrast, RNN-based variants achieve shorter episode lengths of 115 and 116.688, respectively. However, these apparent gains are misleading because they are accompanied by unstable reward learning and collapse into poor FP, which undermines policy reliability. Similarly, the bi-GRU variant produces an average episode length of 114.688 but exhibits high volatility and degraded reward quality, indicating that a shorter episode length alone is not a sufficient indicator of effective learning. 

Table \ref{table7} provides a comparative summary of episode length metrics, including CS, FP, CP, peak episode length (PEL), minimum episode length (MEL), and NF. Overall, the results confirm that the residual-enhanced LSTM architecture not only accelerates the learning of efficient task execution but also maintains stability and robustness. The LSTM-ResB-PPO consistently outperforms other variants by balancing rapid convergence, stable FP and CP, and reliable episode length behavior, while avoiding the instability or collapse observed in alternative encoders. This dual advantage is critical in assistive trajectory optimization, where both timely task completion and consistent behavior across episodes are essential for safety and user trust.

\begin{table*}[htbp]
\centering
\caption{Performance Comparison of Different Algorithms about Mean Ep Length}
\begin{tabular}{lcccccc}
\hline
\textbf{Algorithm} & \makecell{\textbf{CS}} & \makecell{\textbf{FP}} & \makecell{\textbf{CP}} & \textbf{PEL} & \textbf{MEL} & \makecell{\textbf{NF}} \\
\hline
PPO Baseline \cite{6} & 1000000 & 147 & 147 & 147 & 147 & 0 \\
LSTM-ResB PPO & 520000 & 147 & 147 & 147 & 147 & 1 \\
bi-LSTM-ResB PPO & 1560000 & 127.438 & 147 & 147 & 147 & 0 \\
RNN-ResB PPO & 960000 & 115 & 115 & 115 & 115 & 0 \\
bi-RNN-ResB PPO & 1520000 & 116.688 & 114 & 114 & 114 & 2 \\
GRU-ResB PPO & - & 143.188 & - & - & - & 5 \\
bi-GRU-ResB PPO & - & 114.688 & - & - & - & 7 \\
\hline
\end{tabular}
\label{table7}
\end{table*}

\subsection{Cost and Risk Analysis}
For baselines that did not converge within the fixed training budget, we report success and feasibility rates only. To comprehensively assess the impact of MTO and residual-enhanced temporal modeling on strategy safety and cost efficiency, we compare the baseline PPO with the proposed LSTM-ResB-PPO across four indicators: average cost, cost variance, ego risk, and obstacle risk.

\begin{figure*}
\centering
\begin{subfigure}{0.49\linewidth}
    \centerline{\includegraphics[width = .81\textwidth]{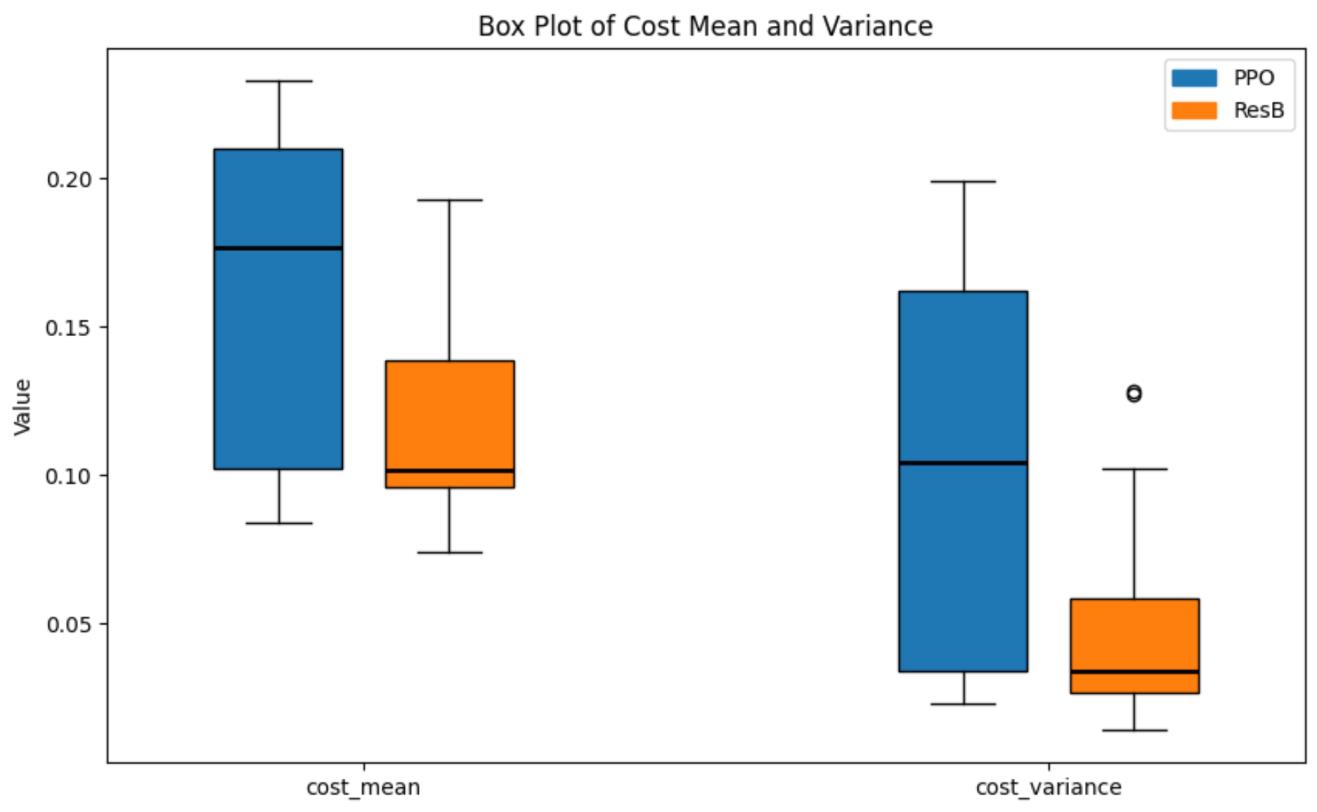}}
    \caption{}
    \label{fig9a}
\end{subfigure}
\begin{subfigure}{0.49\linewidth}
    \centerline{\includegraphics[width = .82\textwidth]{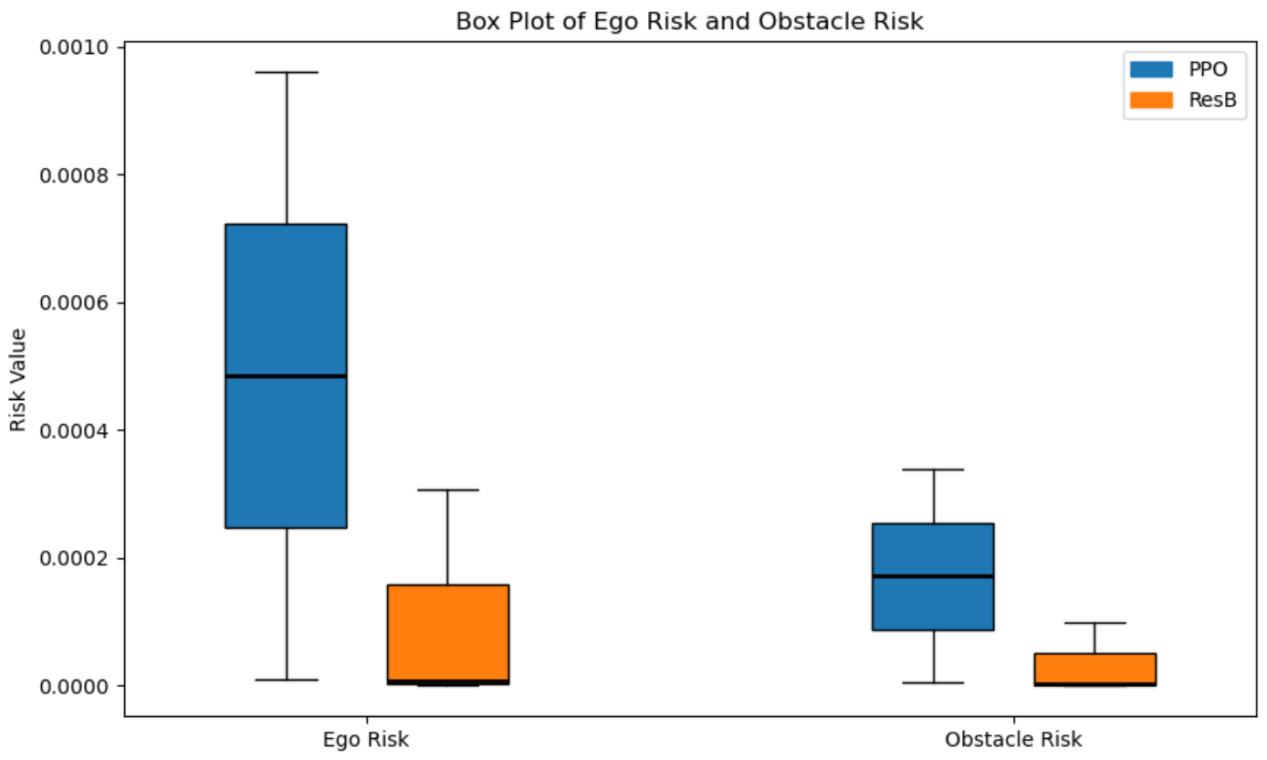}}
    \caption{}
    \label{fig9b}
\end{subfigure}
\caption{Box plots of evaluation metrics for the best models of the two algorithms, where (a) shows cost metrics including mean and variance, and (b) shows risk metrics including mean and variance.
}
\label{fig9}
\end{figure*}

As illustrated in Figure \ref{fig9a}, the LSTM-ResB-PPO achieves a marked reduction in cost. The average cost decreases from 0.166472 in the baseline PPO to 0.116065, a relative reduction of 30.3\%. Meanwhile, the cost variance drops from 0.102921 to 0.048067, denoting a 53.3\% improvement in planning consistency. These improvements occur because MTO suppresses abrupt velocity and acceleration changes, producing smoother trajectories that require fewer corrective maneuvers. Consequently, the agent not only follows more economical paths but also avoids large fluctuations in performance, which directly enhances robustness.

In addition to cost control, Figure \ref{fig9b} highlights significant reductions in risk exposure. The average ego risk decreased to $1.05 \times 10^{-4}$ compared with $4.85 \times 10^{-4}$ for the baseline, representing a 77.2\% reduction. Similarly, the obstacle risk decreased to $3.34 \times 10^{-5}$ compared with $1.72 \times 10^{-4}$, corresponding to an 80.6\% improvement. These results can be attributed to the residual-enhanced LSTM architecture, which captures long-term temporal dependencies and prevents hazardous oscillations in motion decisions. At the same time, MTO optimization enforces kinematic feasibility, steering the agent away from unsafe states and reducing collision likelihood..

Table \ref{table8} summarizes the comparative results, which consistently favor the proposed model. Overall, the LSTM-ResB-PPO generates safer and more economical trajectories than the baseline, achieving stable planning under dynamic conditions. This outcome is particularly relevant for assistive trajectory optimization of visually impaired users, where efficiency and safety must be jointly optimized. The results therefore validate the framework’s ability to address the challenges of multi-objective optimization and robustness identified in the introduction, confirming its practical potential for reliable human-centered trajectory optimization support.
\begin{table*}[htbp]
\centering
\caption{Cost and Risk Metrics for Best Models}
\begin{tabular}{lcccc}
\toprule
\textbf{Algorithm} & \multicolumn{2}{c}{\textbf{Cost}} & \multicolumn{2}{c}{\textbf{Risk}} \\
\cmidrule(lr){2-3} \cmidrule(lr){4-5}
 & \textbf{Avg. Cost-mean} & \textbf{Avg. Cost-variance} & \textbf{Avg. Risk-ego.} & \textbf{Avg. Risk-obs.} \\
\midrule
PPO (Baseline) \cite{6} & 0.166472 & 0.102921 & $4.848913 \times 10^{-4}$ & $1.716621 \times 10^{-4}$ \\
LSTM-ResB-PPO & 0.116065 & 0.048067 & $1.053492 \times 10^{-4}$ & $3.336874 \times 10^{-5}$ \\
\bottomrule
\end{tabular}
\label{table8}
\end{table*}

\subsection{Simulation Results and Quantitative Evaluation}
Building on the scenario selection strategy outlined in the training setup, this section presents a comprehensive evaluation of the proposed MHHTOF using both visualized simulations and quantitative metrics. The aim is to assess whether the framework can reliably meet the planning demands of assistive navigation under realistic, dynamic, and semantically rich driving environments.

Three representative CommonRoad scenarios are employed to ensure robustness and generalizability. These scenarios incorporate dynamic obstacles, semantic elements such as intersections and merging lanes, and moderate complexity in obstacle density and velocity. Their attributes are summarized in Table \ref{table9}, providing transparency and reproducibility. Importantly, these settings approximate the real-world demands of visually impaired scenarios, where multi-agent interactions, semantic awareness, and velocity control are critical.

\begin{table*}
\centering
\caption{Simulation Scenario Properties}
\begin{tabular}{cccc}
\hline
\textbf{Scenarios validation test} & \textbf{DEL\_Lengde-21\_1\_T-15} & \textbf{USA\_Lanker-1\_7\_T-1} & \textbf{ZAM\_Junction-1\_1\_19\_T-1} \\
\hline
Data Source & \makecell{Scenario Factory 2.0 \\- SUMO} & \makecell{Generation (NGSIM), \\OSM} & \makecell{Specification-based\\ scenario synthesis} \\
Obstacle Behavior Type & TRAJECTORY & TRAJECTORY & TRAJECTORY \\
Scenario Tags & \makecell{INTERSECTION, \\MERGING LANES, \\SIMULATED,\\ SINGLE LANE}&\makecell{ MULTI LANE, \\SPEED LIMIT,\\ ONCOMING TRAFFIC, \\ COMFORT, \\URBAN, \\INTERSECTION,\\ LANE FOLLOWING} & \makecell{TWO LANE, \\ONCOMING TRAFFIC, \\TURN LEFT, \\URBAN,\\ INTERSECTION} \\
Obstacle Classes & CAR & CAR & CAR \\
Goal Types & \makecell{POSITION,\\ TIME\_STEP} & \makecell{ORIENTATION,\\ VELOCITY,\\ TIME\_STEP} & \makecell{LANEFT,\\ VELOCITY,\\ TIME\_STEP} \\
Scenario Duration & 15.10 s & 1.50 s & 14.70 s \\
Initial Velocity & 5.11 m/s & 9.73 m/s & 5.16 m/s \\
No. of Static Obstacles & 0 & 0 & 0 \\
No. of Dynamic Obstacles & 89 & 22 & 5 \\
No. of Ego Vehicles & 1 & 1 & 1 \\
\hline
\end{tabular}
\label{table9}
\end{table*}
To illustrate planning dynamics in greater detail, two scenarios are selected for trajectory-level visualization. For each case, four sets of indicators are analyzed: (i) global path execution, (ii) velocity and acceleration profiles, (iii) cumulative weighted planning cost, and (iv) cumulative change in predicted actions, which serve as a proxy for temporal consistency and decision reliability.

\subsubsection{Case Study 1: DEU\_Lengede-21\_1\_T-15}
\begin{figure*}
\centering
\begin{subfigure}[c]{0.7\linewidth}
    \centerline{\includegraphics[width = .89\textwidth]{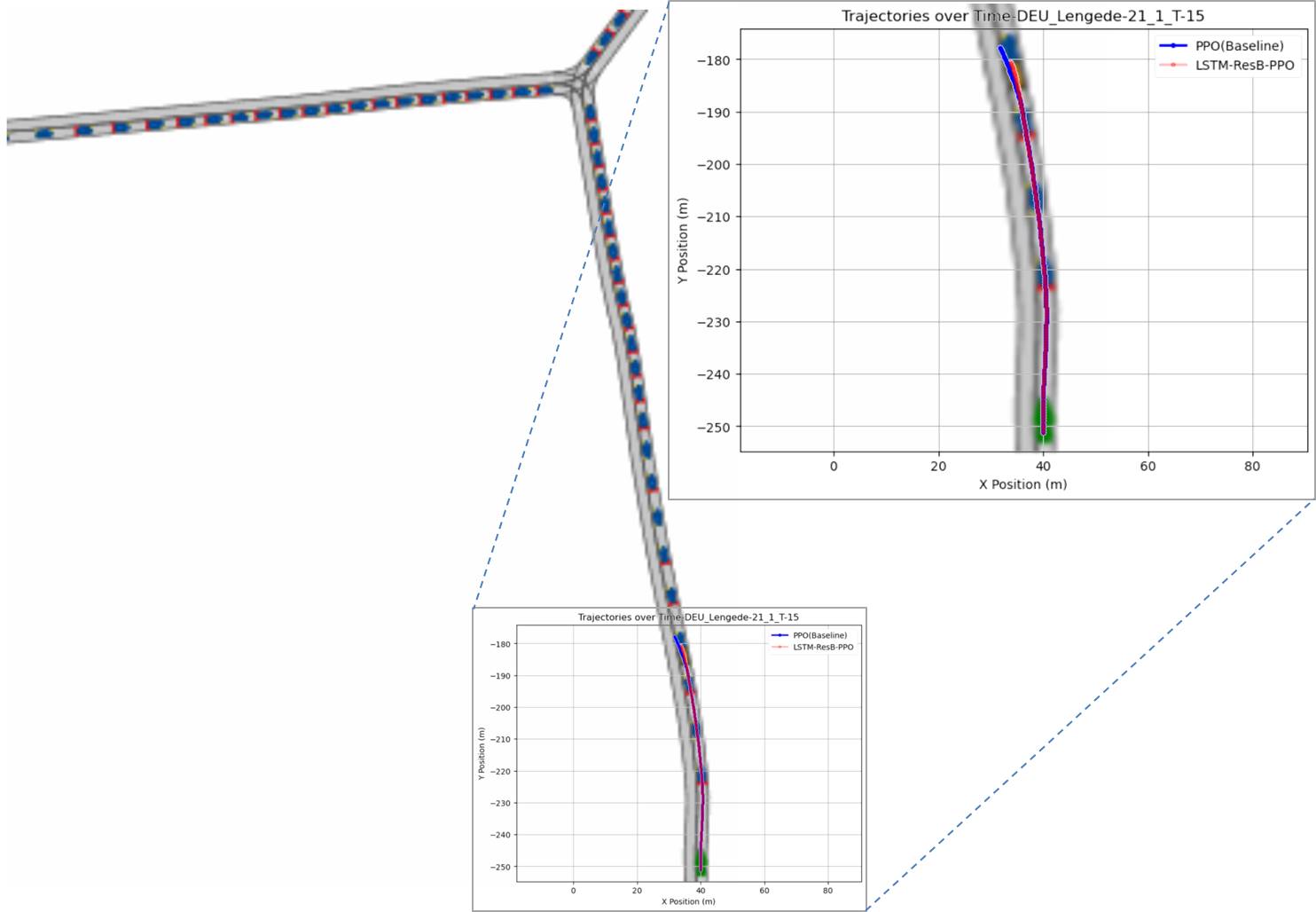}}
    \caption{}
    \label{fig10a}
\end{subfigure}

\begin{subfigure}[c]{0.7\linewidth}
    \centerline{\includegraphics[width = .89\textwidth]{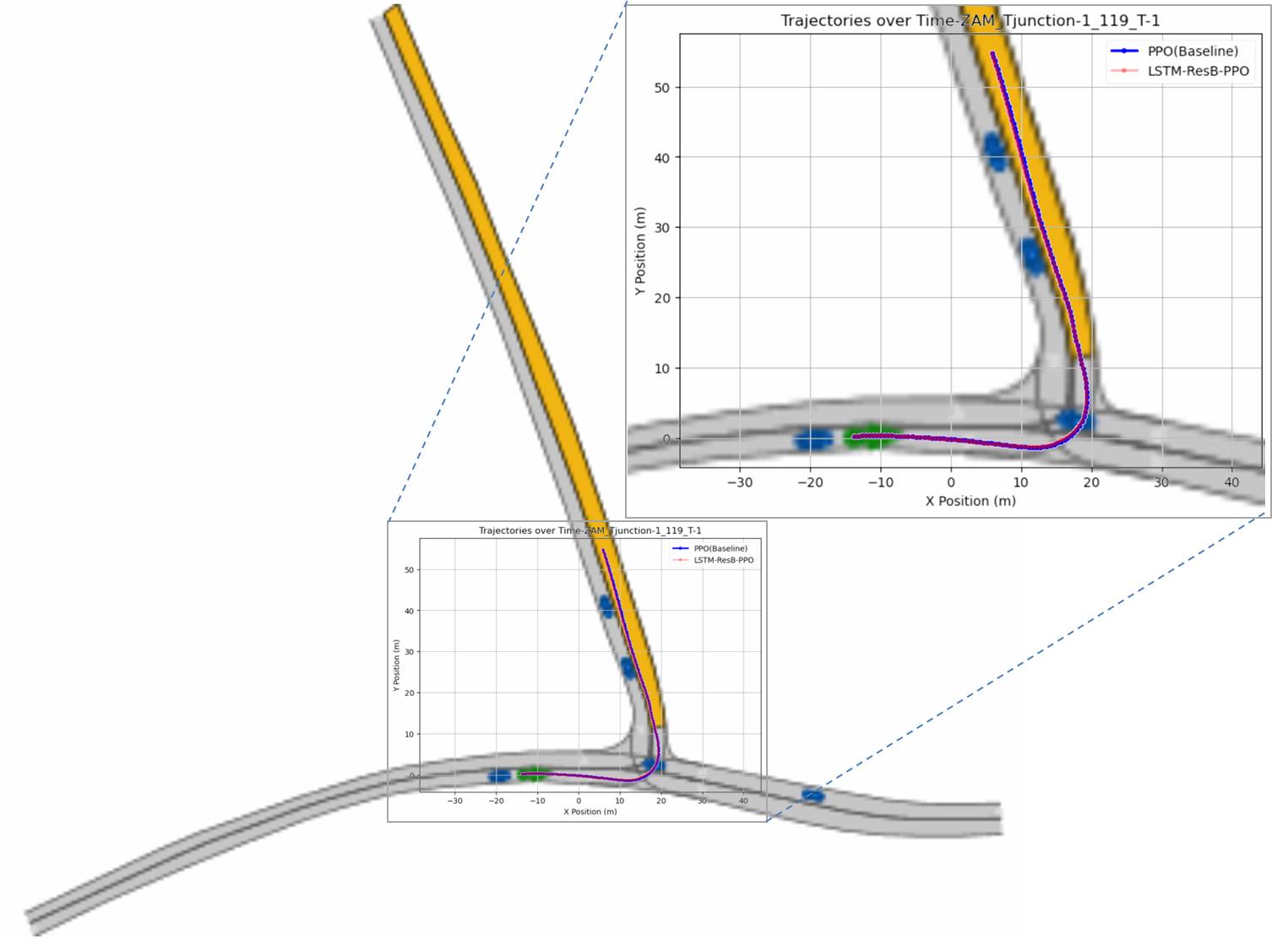}}
    \caption{}
    \label{fig10b}
\end{subfigure}
\caption{Visualization results of trajectory planning in two CommonRoad scenarios, where (a) DEU\_Lengede-21\_1\_T-15 shows the algorithmic simulation outcomes and (b) ZAM\_Junction-1\_119\_T-1 illustrates comparative trajectory planning results.
}
\label{fig10}
\end{figure*}

\begin{figure*}
\centering
\begin{subfigure}[c]{.49\linewidth}
    \centerline{\includegraphics[width = .99\textwidth]{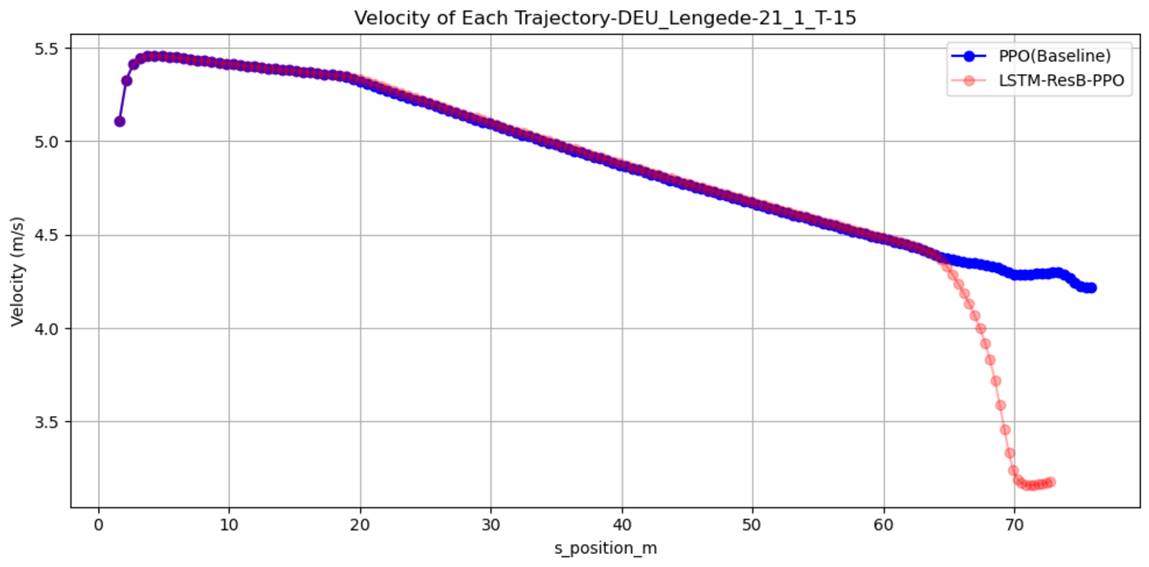}}
    \caption{}
    \label{fig11a}
\end{subfigure}
\begin{subfigure}[c]{.49\linewidth}
    \centerline{\includegraphics[width = .99\textwidth]{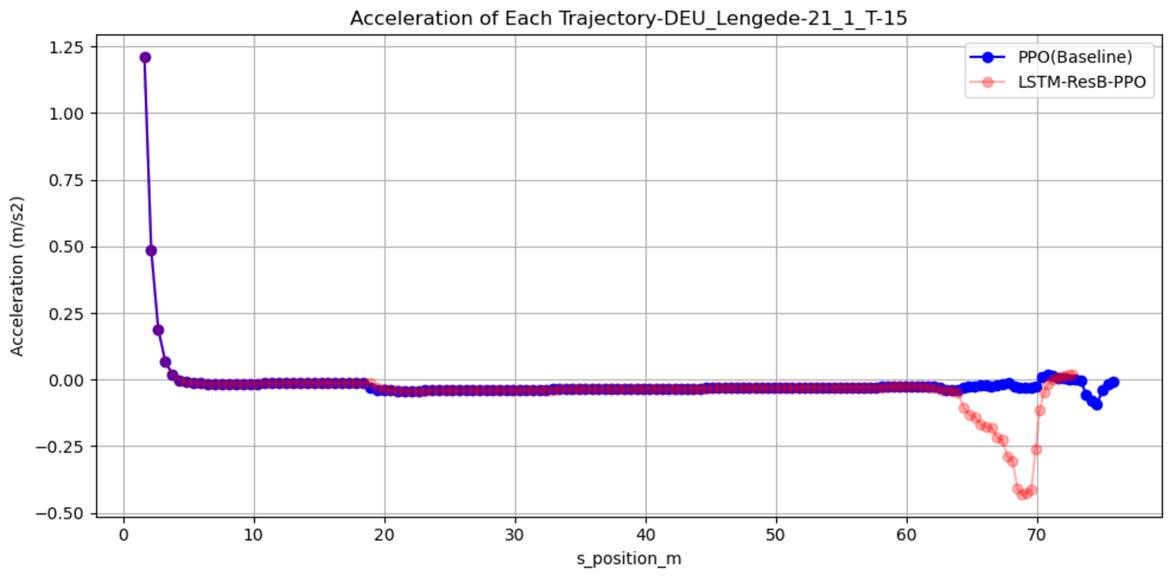}}
    \caption{}
    \label{fig11b}
\end{subfigure}
\begin{subfigure}[c]{.49\linewidth}
    \centerline{\includegraphics[width = .99\textwidth]{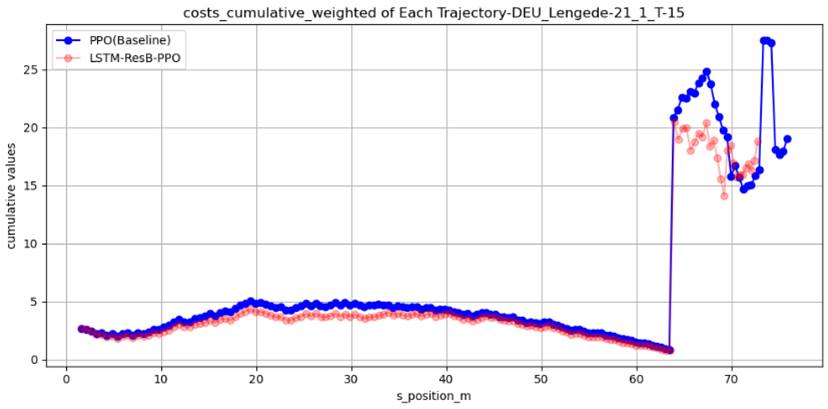}}
    \caption{}
    \label{fig11c}
\end{subfigure}
\caption{Comparison of two algorithms under the DEU\_Lengede-21\_1\_T-15 scenario, where (a) speed curves along the S-direction position, (b) acceleration curves along the S-direction position, and (c) line comparison of cumulative weighted cost are illustrated.
}
\label{fig11}
\end{figure*}

As shown in Figure \ref{fig10a}, a marked performance gap emerges between the baseline PPO and the proposed framework. The baseline model fails to complete the navigation task and deviates into an oncoming traffic lane, generating hazardous behavior and violating traffic semantics. In contrast, the LSTM-ResB-PPO adheres to lane topology, completing the task without violation, thereby evidencing its superior understanding of interaction and environmental constraints.
The velocity and acceleration profiles shown in Figure \ref{fig11a} and Figure \ref{fig11b} further highlight this distinction. The baseline exhibits abrupt fluctuations, while the proposed model achieves a gradual, well-controlled velocity decline. This smooth deceleration not only reflects kinematic feasibility but also aligns with the comfort and safety requirements of assistive navigation for visually impaired users.

\begin{figure*}
\centering
\begin{subfigure}[c]{.49\linewidth}
    \centerline{\includegraphics[width = .95\textwidth]{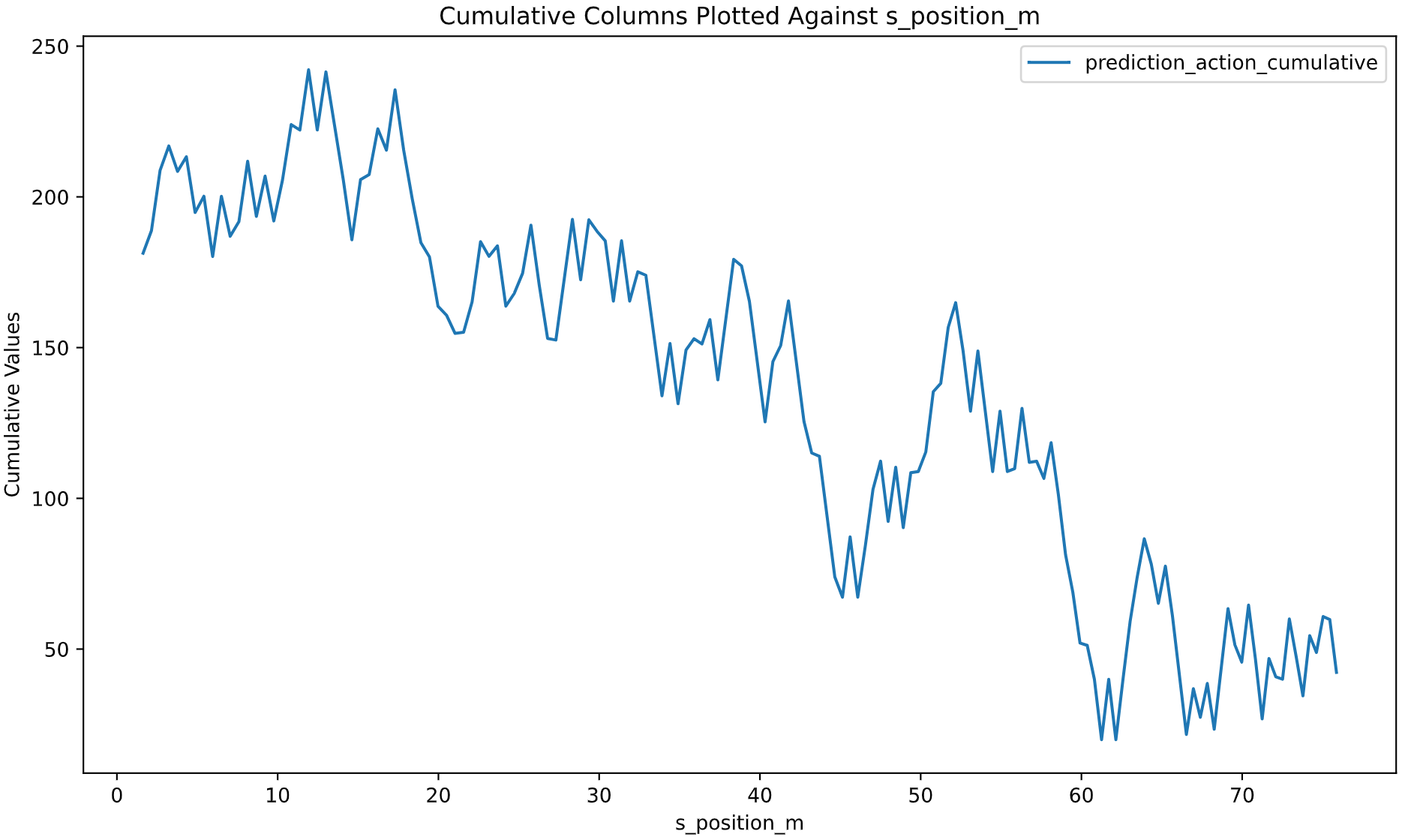}}
    \caption{}
    \label{fig12a}
\end{subfigure}
\begin{subfigure}[c]{.49\linewidth}
    \centerline{\includegraphics[width = .95\textwidth]{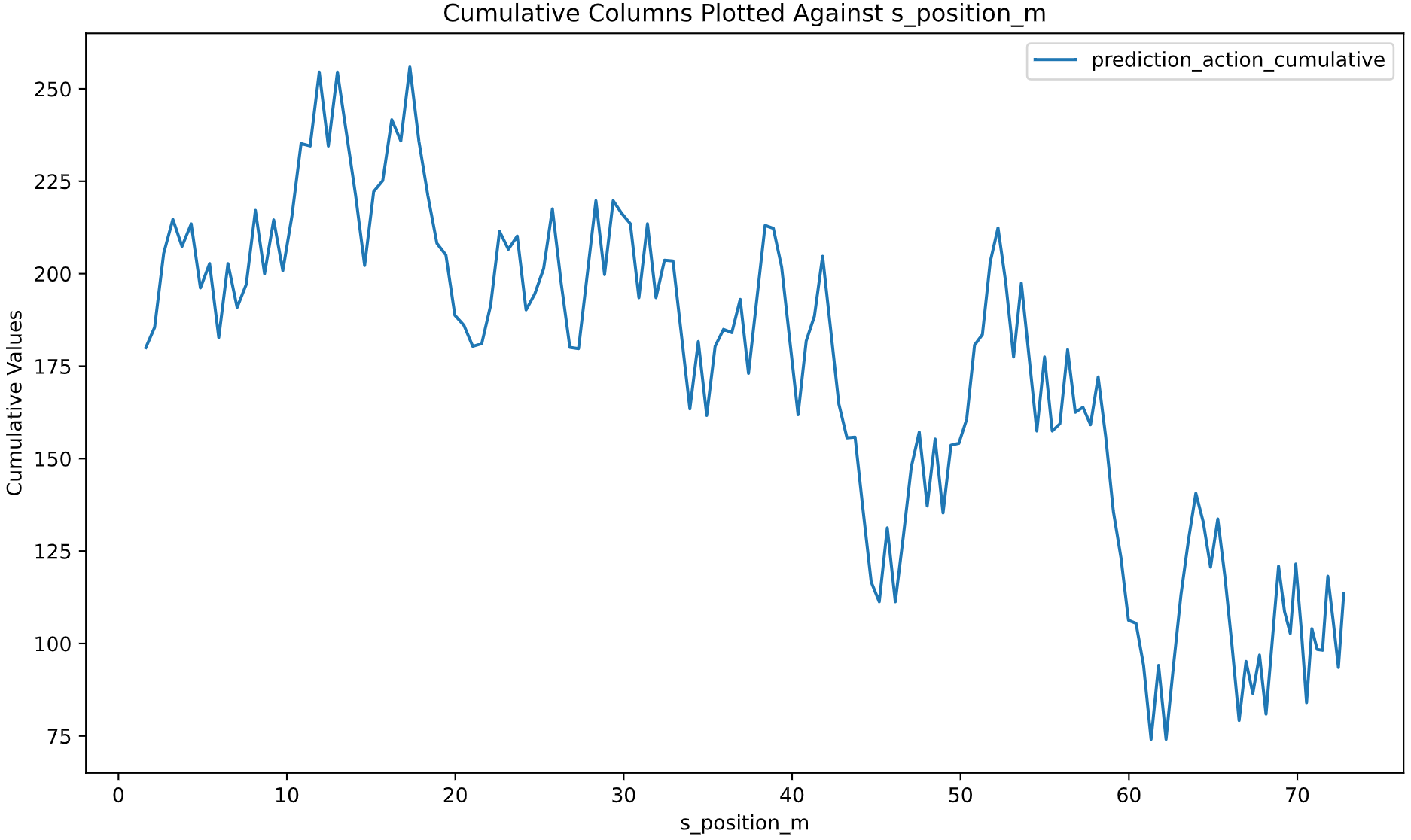}}
    \caption{}
    \label{fig12b}
\end{subfigure}
\caption{Variation of cumulative weighted cost and cumulative predicted action with s-direction position for two algorithms under the DEU\_Lengede-21\_1\_T-15 Scenario: (a) line variation of cumulative predicted action for LSTM-ResB-PPO; (b) line variation of cumulative predicted action for the baseline algorithm.}
\label{fig12}
\end{figure*}

From the optimization perspective, the cumulative weighted cost shown in Figure \ref{fig11c} remains consistently lower for LSTM-ResB-PPO, demonstrating more efficient trajectory selection under multi-objective trade-offs. Furthermore, the cumulative predicted action deviation shown in Figures \ref{fig12} reveals that the baseline suffers from increasing instability, whereas the proposed model converges toward a stable downward trend, confirming its temporal consistency and robustness.

Overall, this scenario shows that residual-enhanced temporal modeling enables the proposed framework to maintain semantic correctness, motion smoothness, and decision consistency, thereby avoiding the collapse behaviors observed in the baseline.

\subsubsection{Case Study 2: ZAM\_Junction-1\_119\_T-1}
\begin{figure*}
\centering
\begin{subfigure}[c]{.49\linewidth}
    \centerline{\includegraphics[width = .99\textwidth]{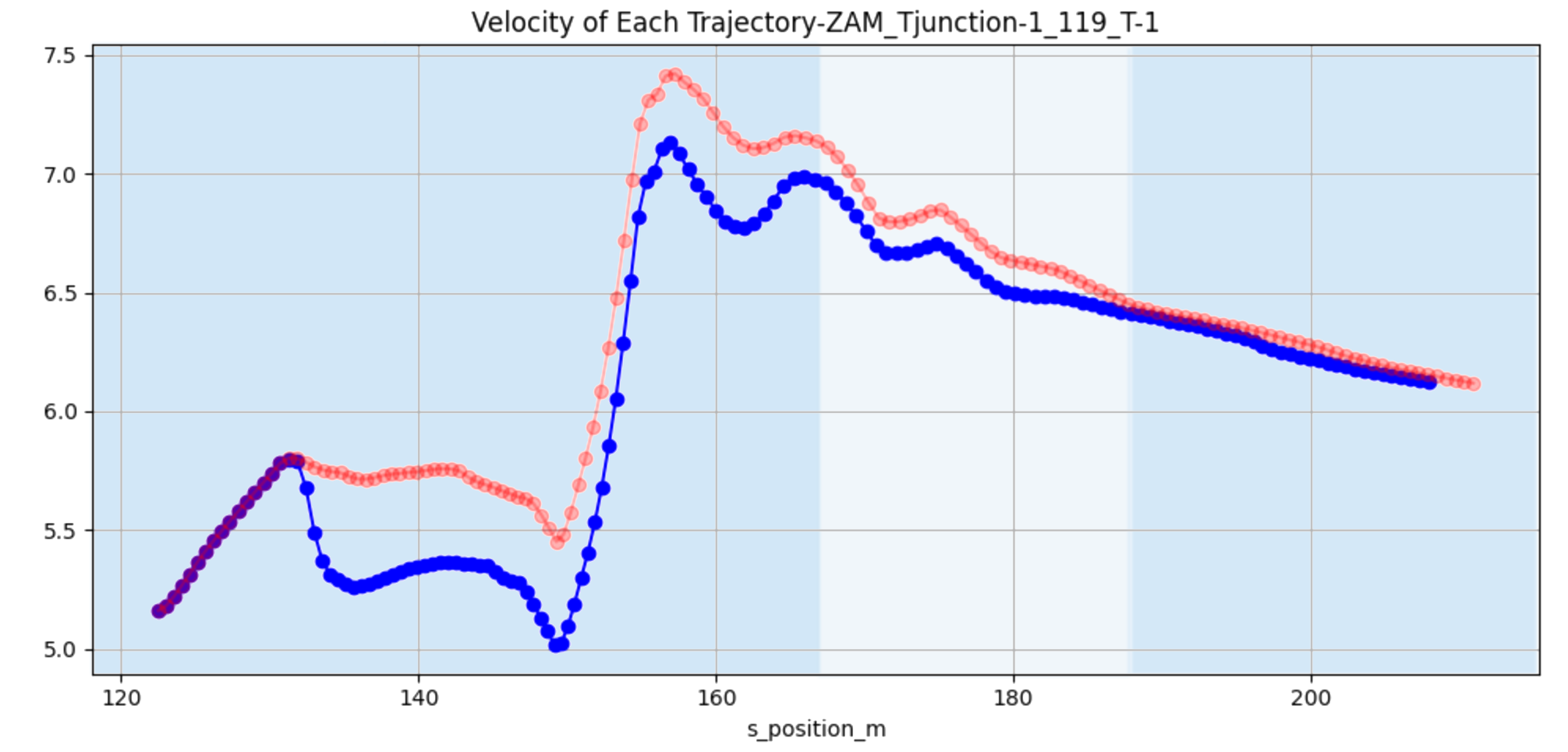}}
    \caption{}
    \label{fig13a}
\end{subfigure}
\begin{subfigure}[c]{.49\linewidth}
    \centerline{\includegraphics[width = .99\textwidth]{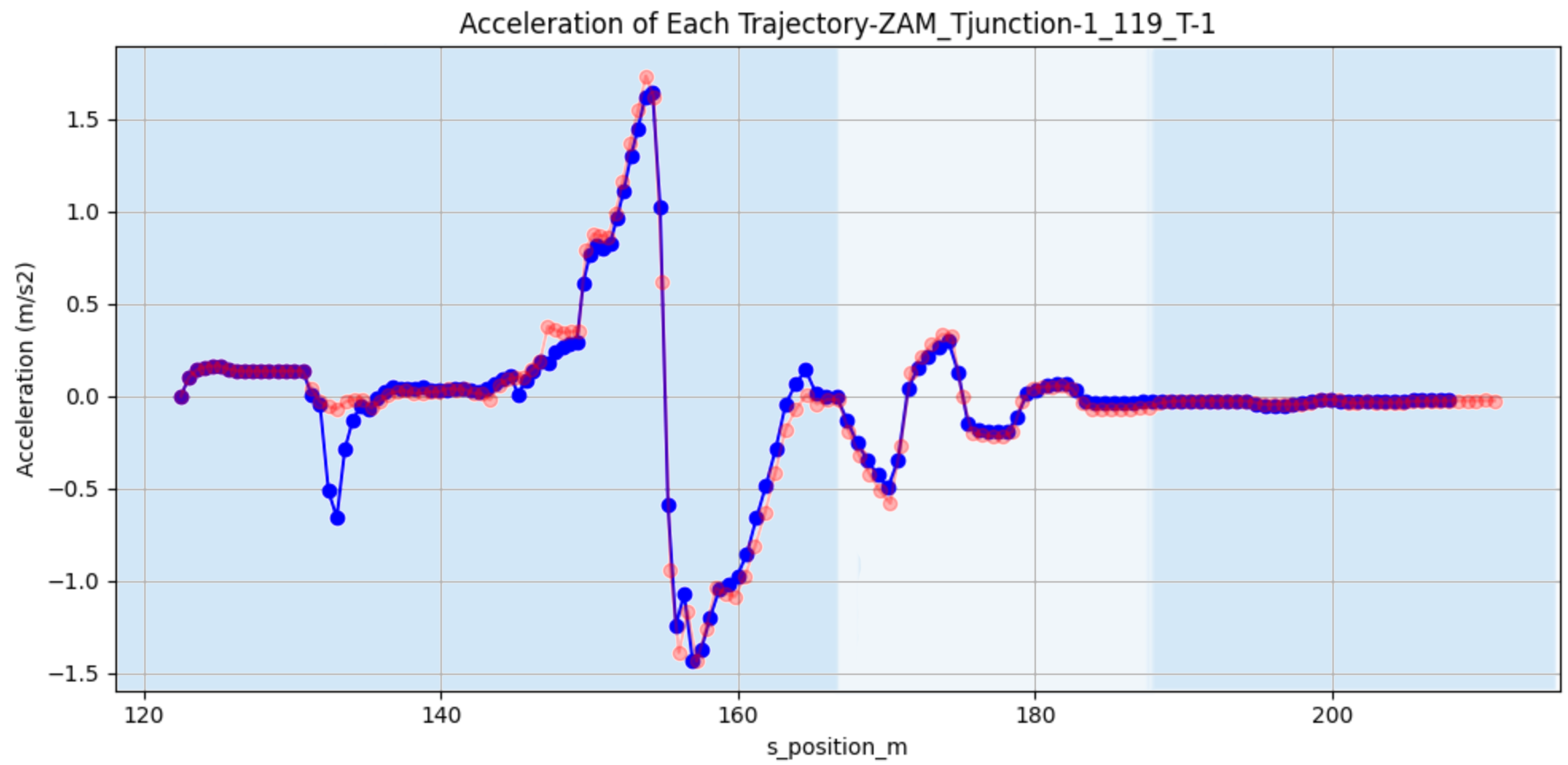}}
    \caption{}
    \label{fig13b}
\end{subfigure}
\begin{subfigure}[c]{.49\linewidth}
    \centerline{\includegraphics[width = .99\textwidth]{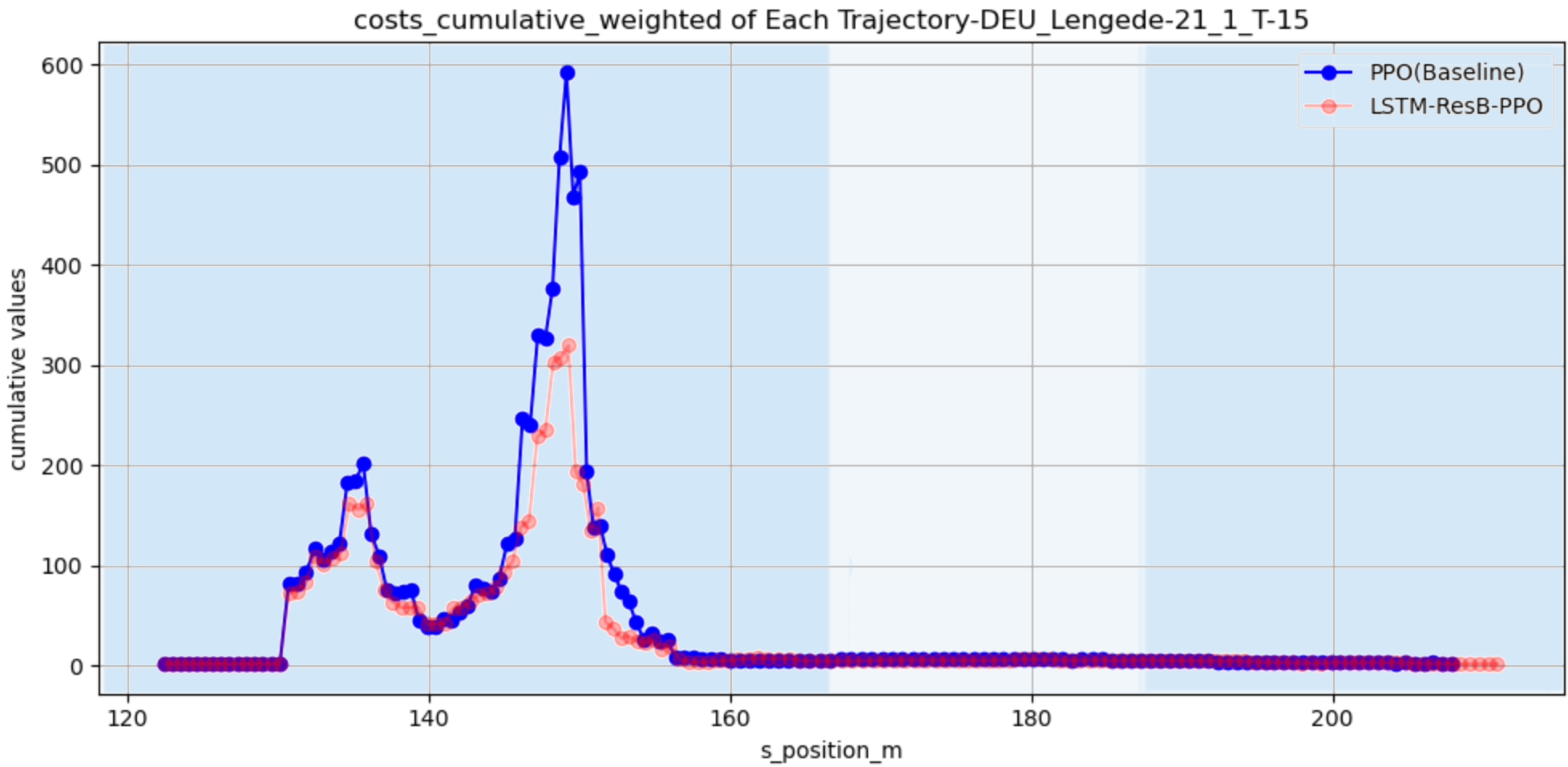}}
    \caption{}
    \label{fig13c}
\end{subfigure}
\caption{Comparison of two algorithms under the ZAM\_Junction-1\_119\_T-1 scenario, where (a) speed curves along the S-direction position, (b) acceleration curves along the S-direction position, and (c) line comparison of cumulative weighted cost are illustrated.
}
\label{fig13}
\end{figure*}

In this scenario shown in Figure \ref{fig10b}, both models satisfy the predefined goal conditions. However, the quality of motion differs substantially. The LSTM-ResB-PPO agent follows a slightly longer path but benefits from smoother control transitions, as evidenced by the velocity and acceleration curves shown in Figure \ref{fig13a} and \ref{fig13b}. Its trajectory naturally divides into three interpretable phases: left-turn interaction, deceleration, and stabilization. Each phase is characterized by more continuous and refined control compared to the baseline.

During the left-turn maneuver, where the agent must yield to oncoming traffic, the baseline produces abrupt peaks in velocity and acceleration, while the proposed framework maintains stability, mitigating safety risks. In the deceleration phase, LSTM-ResB-PPO gradually reduces velocity without sharp braking, avoiding comfort-reducing oscillations. Finally, in the stabilization phase, the framework gently converges toward the target velocity, demonstrating fine-grained low-speed regulation essential for human-centered assistive planning.

\begin{figure*}
\centering
\begin{subfigure}[c]{.49\linewidth}
    \centerline{\includegraphics[width = .95\textwidth]{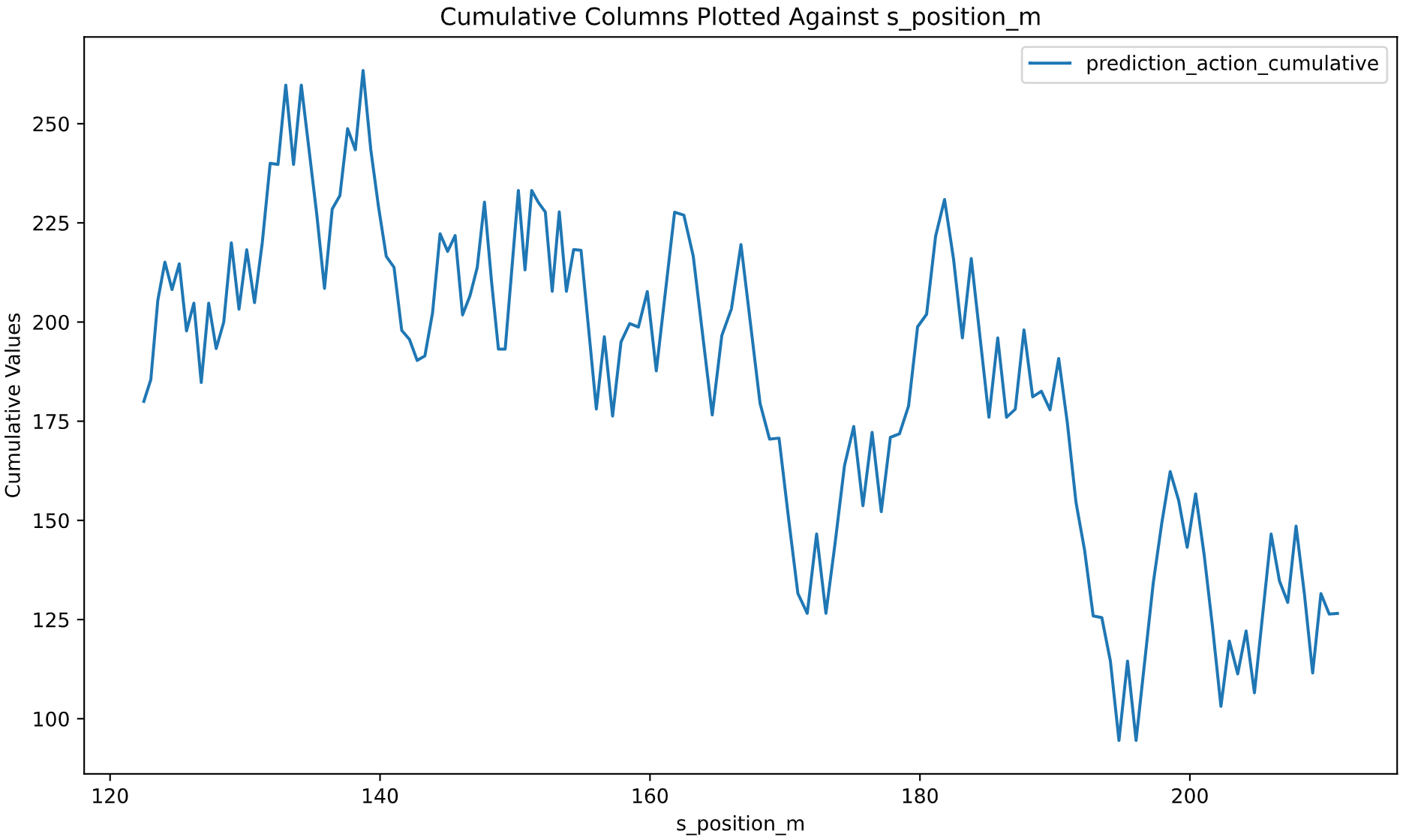}}
    \caption{}
    \label{fig14a}
\end{subfigure}
\begin{subfigure}[c]{.49\linewidth}
    \centerline{\includegraphics[width = .95\textwidth]{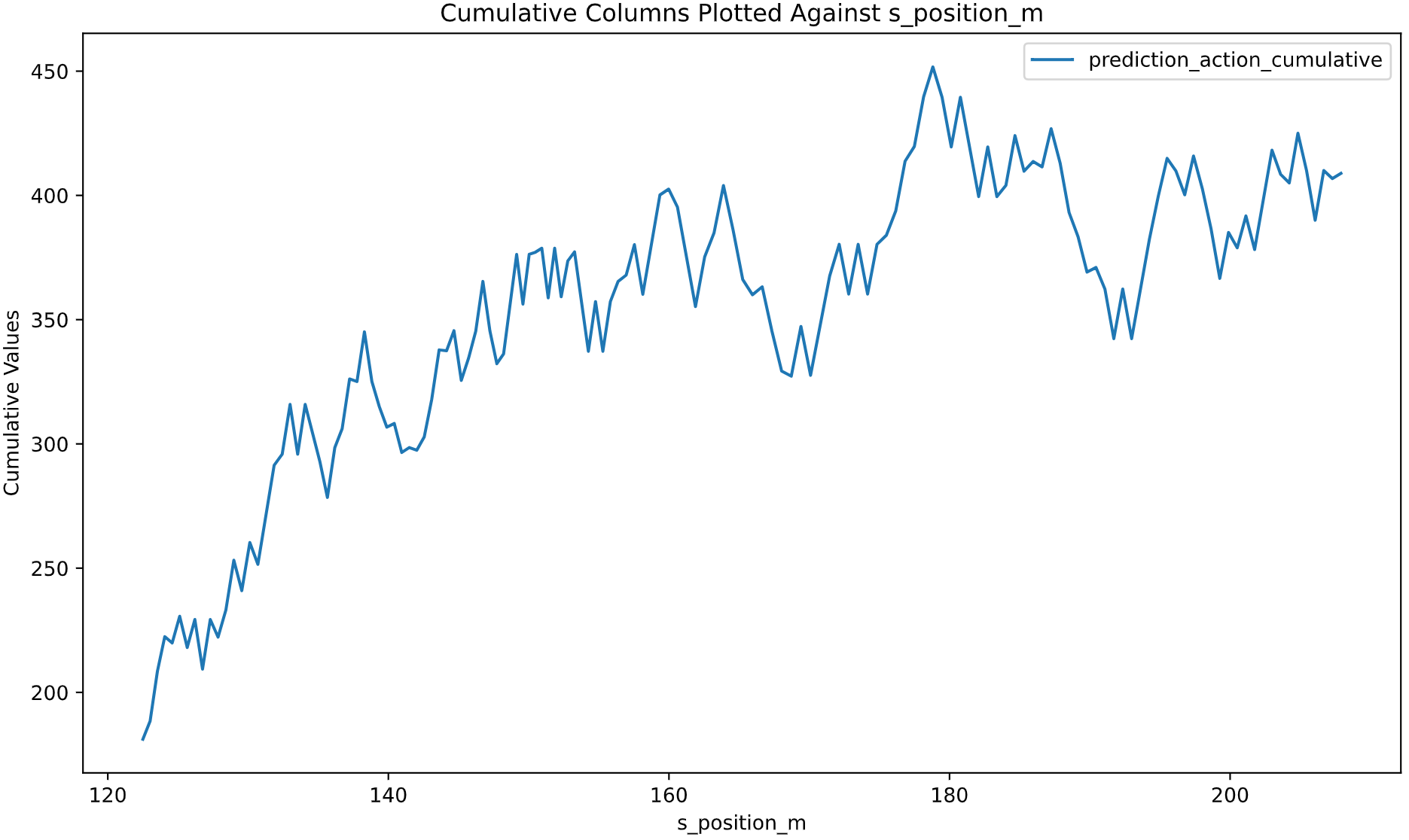}}
    \caption{}
    \label{fig14b}
\end{subfigure}
\caption{Variation of cumulative weighted cost and cumulative predicted action with s-direction position for two algorithms under the ZAM\_Tjunction-1\_119\_T-1 scenario: (a) line variation of cumulative predicted action for LSTM-ResB-PPO; (b) line variation of cumulative predicted action for the baseline algorithm.}
\label{fig14}
\end{figure*}

The cost-related indicators shown in Figure \ref{fig13c} and \ref{fig14} reinforce these observations. The cumulative weighted cost remains lower and smoother for the proposed framework, reflecting its ability to optimize multi-objective trade-offs across the planning horizon. Meanwhile, the cumulative predicted action deviation shows a more monotonic decline, indicating reduced planning uncertainty and enhanced risk awareness. The increasing trend indicates unstable decision shifts while a decreasing trend reflects convergence toward stable and reliable planning.

Overall, this case highlights that LSTM-ResB-PPO not only achieves task success but also ensures higher-order motion qualities such as smoothness, stability, and low uncertainty, which are critical for trustworthy assistive trajectory optimization.

\subsubsection{Cross-Scenario Evaluation}
Across all three scenarios, the LSTM-ResB-PPO achieves full task completion, while the baseline fails in high-risk settings. The proposed model consistently generates smoother velocity and acceleration profiles, maintains convergent and moderate cumulative costs, and produces action predictions with lower volatility. These findings collectively underscore its ability to deliver safe, efficient, and temporally consistent trajectories under diverse and dynamic conditions.

Table \ref{table10} quantitatively substantiates these advantages. Compared to the baseline, the LSTM-ResB-PPO achieves equal or higher goal success rates, reduces trajectory cluster counts indicating greater planning efficiency, significantly lowers ego-risk measures, and consistently achieves lower cumulative weighted costs. These improvements can be attributed to the synergistic integration of residual-enhanced feature representation and LSTM-based temporal modeling, which together strengthen both short-term motion control and long-term decision optimization.

\begin{table*}[htbp]
\centering
\caption{Performance Comparison of Algorithms Across Different Scenarios}
\begin{tabular}{lccccc}
\toprule
\textbf{Scenarios} & \textbf{Algorithm} & \textbf{Goal Reached} & \makecell{\textbf{Number of}\\ \textbf{Trajectory Clusters}} & \textbf{Avg. Ego-risk} & \textbf{Avg. Costs} \\
\midrule
DEU\_Lengede-21\_1\_T-15 & PPO (Baseline) \cite{6} & 0 & 152 & 2.25395e-14 & 6.6384 \\
& LSTM-ResB-PPO & 1 & 150 & 8.4692e-16 & 5.4286 \\[3pt]
ZAM\_Junction-1\_119\_T-1 & PPO (Baseline) \cite{6} & 1 & 146 & 9.7826e-05 & 53.015 \\
& LSTM-ResB-PPO & 1 & 146 & 8.9332e-05 & 36.3689 \\[3pt]
USA\_Tanker-1\_7\_T-1 & PPO (Baseline) \cite{6} & 1 & 13 & 3.2864e-05 & 31.5190 \\
& LSTM-ResB-PPO & 1 & 13 & 3.2864e-05 & 30.6837 \\
\bottomrule
\end{tabular}
\label{table10}
\end{table*}

\subsubsection{Final Summary}
Taken together, the results demonstrate that the proposed MHHTOF not only ensures feasibility and goal achievement but also advances critical aspects of assistive trajectory optimization, including safety, smoothness, stability, and cost-efficiency. By avoiding collapse behaviors, reducing abrupt dynamics, and enhancing temporal reliability, the framework addresses core challenges of multi-objective trajectory optimization in visually impaired scenarios.

\subsection{Discussion}
These findings collectively validate the efficacy of the proposed trajectory optimization framework in assistive navigation contexts. The integration of LSTM-based temporal modeling and residual blocks not only improves policy stability and behavioral smoothness but also enables a favorable trade-off between exploration efficiency and conservative decision-making.  Coupled with MTO constraints implicitly embedded in the DRL structure, the model exhibits strong adaptability across diverse scenes involving uncertainty, interaction, and physical limitations.

Overall, the simulation-based analysis confirms that LSTM-ResB-PPO offers a reliable and scalable solution for safety-critical, comfort-sensitive planning tasks, establishing a solid foundation for downstream deployment in real-world assistive trajectory optimization framework.
\section{Conclusion}
\label{c}
This paper presents a human-centered MHHTOF framework that integrates HTSCMOE combining HTSC and MTO with residual-enhanced DRL, thereby enabling semantically aligned and dynamically feasible trajectory planning for assistive trajectory optimization. Furthermore, the incorporation of temporal modeling and residual connections strengthens policy stability and enhances representational capacity. By introducing DCMM across Frenet and Cartesian stages, the framework ensures consistent optimization while capturing personalized requirements of visually impaired users. 

Extensive simulations across diverse CommonRoad scenarios validate the effectiveness of the proposed LSTM-ResB-PPO policy compared with PPO baselines. The results confirm superior convergence behavior, improved reward attainment, and significant reductions in cost and risk indicators. These improvements highlight the framework’s advantages in safety-critical and comfort-sensitive planning tasks, particularly in complex urban intersections and dynamic obstacle environments.

Despite these promising outcomes, the current framework is limited to two-dimensional navigation settings and does not explicitly address scalability to large-scale real-world deployments. Future research could extend the approach to 3D navigation tasks, incorporate richer sensory modalities, and further optimize temporal decision-making in highly dynamic and uncertain environments. In addition, investigating the integration of real-world user feedback into the cost modeling process may enhance personalization and practical feasibility in assistive navigation applications.

\bibliographystyle{elsarticle-num}
\bibliography{ref}

\end{document}